\documentclass[11pt]{article}

\usepackage[final]{acl}

\usepackage{times}
\usepackage{bm}
\usepackage{latexsym}
\usepackage{booktabs}
\usepackage{float}
\usepackage[T1]{fontenc}

\usepackage[utf8]{inputenc}

\usepackage{microtype}

\usepackage{inconsolata}

\usepackage{graphicx}
\usepackage{subcaption}
\usepackage{array, makecell} 
\usepackage{xcolor}
\usepackage{amsmath}

\title{Imag-Eval: a language-grounded framework for interpretable Text-to-Image instruction following evaluation \thanks{This paper is appearing in the Proceedings og the 2026 Conference of Empirical Methods in Natural Language Processing (EMNLP 2026). Please cite the EMNLP version.}}

\author{Ibrahim MOHAMED SEROUIS \thanks{Corresponding author, main contributor.} \\
  Talan Research and Innovation Center \\ 
  Toulouse, France \\
  \texttt{ibrahim.mohamed-serouis@talan.com} \\\And
  David JARAMILLO DUQUE  \\
  Talan Research and Innovation Center \\ 
  Paris, France \\
  \texttt{david.jaramillo-duque@talan.com} \\}

\begin{document}
\maketitle
\begin{abstract}
Text-to-Image (T2I) models have recently achieved impressive visual fidelity, yet their evaluation remains constrained by benchmarks that are often difficult to interpret and insufficiently diagnostic. Existing skill-based evaluations tend to overlook critical failure modes that strongly impact usability but fall outside standard taxonomies, such as global incoherence arising from missing parts or physically implausible configurations (e.g., floating objects). In addition, prompt difficulty is typically controlled along a single dimension; either prompt length or the number of elements to generate. To address these limitations, we introduce \textsc{Imag-Eval}, a controlled benchmark designed to assess how T2I models ground compositional natural-language instructions into visual outputs. Unlike prior work that conflates surface linguistic complexity with compositional difficulty, \textsc{Imag-Eval} explicitly seeks to disentangle these factors by independently varying both the number of instances and the combination of constraints (rules), while avoiding error propagation. This design enables fine-grained and interpretable analysis of \emph{where} cross-modal instruction following fails. Our benchmark comprises 1{,}140 prompts and 8{,}842 combined rules, and we evaluate it on several state-of-the-art models. Complementing this analysis with an additional study of over 2{,}000 prompts from a concurrent benchmark, our results suggest that, for structured skills, compositional difficulty is primarily governed by the number of grounded rules and their binding to instances, rather than by prompt length alone.
Code samples are available here: \url{https://github.com/Justsecret123/Imag-Eval}
\end{abstract}

\section{Introduction}

Text-to-Image generation has advanced rapidly, with modern systems capable of producing visually realistic and diverse images from natural language prompts, spanning a wide range of styles \cite{yu_multi-style_2024}, lighting conditions, and even multilingual rendering. Despite these improvements, reliable and interpretable evaluation remains a bottleneck.

Early evaluation approaches relied on embedding-based metrics such as CLIPScore \cite{hessel_clipscore_2021}, which approximate text–image alignment but provide limited diagnostic insight. More recent benchmarks move toward skill-based evaluation, decomposing performance into capabilities across different difficulty levels. However, these approaches remain limited in interpretability. First, they often fail to capture whether all required entities are generated coherently, overlooking critical failure modes such as missing parts or globally inconsistent scenes. Second, prompt difficulty is typically controlled along a single axis, most commonly prompt length, or per-skill hardness without explicitly accounting for the underlying compositional structure of the instruction. Third, evaluation pipelines frequently propagate upstream errors (e.g., missing objects) into downstream skill failures, leading to double penalization and obscuring the source of model breakdowns. To address these limitations, we introduce:

\emph{\textbf{(1) Imag-Eval}}, a controlled evaluation framework designed to provide fine-grained and interpretable analysis of compositional instruction following in T2I models. Unlike prior approaches that rely on proxy measures such as prompt length, Imag-Eval explicitly seeks to disentangle surface linguistic complexity from compositional difficulty by varying two orthogonal factors that we name \emph{compositional load}: the number of instances and the combination of constraints. This design enables us to directly investigate where models fail as compositional load increases. Furthermore, we introduce an under-explored yet critical evaluation dimension, coherent and complete object generation.

\emph{\textbf{(2) A benchmark dataset supporting this framework}}, comprising 1{,}140 prompts and 8{,}842 evaluation rules. We validate the proposed methodology through extensive experiments involving 14 annotators, over 8{,}000 annotations, and more than 6{,}000 generated images evaluated across a diverse set of proprietary and open-source models.

Our empirical findings, including an additional analysis of more than 2{,}000 prompts from a concurrent benchmark, provide broader insights into current evaluation practices. In particular, we show that model performance is primarily driven by \emph{compositional load} (the number of grounded rules and their binding to instances) rather than by surface linguistic properties such as prompt length, for structured skills. This highlights the importance of disentangling linguistic complexity from compositional structure, and motivates the need for controlled benchmarks that enable more interpretable and diagnostic assessment of T2I models.

\section{Related Work}

\textbf{Interpretable/Explainable AI (XAI)} aims to make model behavior and failure modes understandable, enabling more reliable diagnosis and comparison beyond aggregate metrics \cite{ribeiro_2016,doshi-velez_towards_2017,lipton2018mythos}. A key principle is that evaluation should be \emph{diagnostic}: rather than reporting a single scalar score, it should decompose performance into meaningful dimensions that help localize errors. This requirement is particularly critical for generative multimodal systems, where high-level similarity metrics (e.g., CLIP-based scores \cite{hessel_clipscore_2021}) can conflate distinct failure sources, such as missing entities and incorrect bindings. In response, recent work advocates skill-based evaluations that assess models along semantically grounded axes \cite{ribeiro_beyond_2020,srivastava_beyond_2023,afkanpour_automated_2025}.

\textbf{Image generation.} Image generation has evolved from generative adversarial networks (GANs) \cite{goodfellow_generative_2014} and variational autoencoders (VAEs) \cite{kingma_auto-encoding_2013} to diffusion-based models \cite{sohl-dickstein_deep_2015,ho_denoising_2020}, which now dominate the field. Diffusion models achieve strong fidelity and diversity through iterative denoising and underpin many modern T2I systems \cite{rombach_high-resolution_2022}. More recent architectures (e.g., Stable Diffusion XL, Stable Cascade, Qwen-Image) further extend these capabilities, improving both visual quality and controllability \cite{podell_sdxl_2024,pernias_wurstchen_2024,wu_qwen-image_2025}. Despite these advances, evaluation has not kept pace, particularly in compositional instruction following.

\textbf{Text-to-Image benchmarks.} Early multi-task T2I benchmarks defined broad evaluation categories, sometimes with difficulty tiers, but typically assessed skills in isolation, limiting the study of \emph{compositional} failures across interacting constraints \cite{petsiuk_human_2022}. DALL-Eval \cite{cho_dall-eval_2023} introduced automated evaluation for a limited set of skills (e.g., counting, spatial relations), but provides limited coverage of global coherence issues. More recent benchmarks incorporate multiple constraints within single prompts. HRS-Bench \cite{bakr_hrs-bench_2023} expands coverage to diverse attributes, including emotion and robustness, yet its notion of difficulty is primarily rule-centric. Multi-skill evaluations such as T2I-CompBench(++) \cite{huang_t2i-compbench_2023,huang_t2i-compbench_2025} assess attribute binding and object relations across curated tasks, but do not explicitly control difficulty across multiple interacting factors.

TIIF-Bench \cite{wei_tiif-bench_2025} highlights the role of prompt length by comparing equivalent short and long prompts, demonstrating that length can confound evaluation. However, as suggested in our analysis, prompt length alone is an incomplete proxy for difficulty, as it may vary independently of the underlying compositional structure. This observation motivates the need to disentangle surface linguistic properties from compositional load when assessing instruction-following capabilities.

Across benchmarks, an additional limitation is that evaluation pipelines often propagate upstream errors (e.g., missing objects) into downstream skill failures, effectively double-counting errors and obscuring their origin. Moreover, global coherence issues such as missing parts or physically implausible configurations remain underrepresented in standard taxonomies. While specialized efforts target related artifacts (e.g., human-body realism), they do not provide a general, model-agnostic framework for evaluating coherence across all entities \cite{andreou_bodymetric_2024,corneanu_structured_2025}.

\textbf{Positioning of Imag-Eval.} \textsc{Imag-Eval} builds on these lines of work by introducing a controlled and explicitly factorized evaluation framework. In contrast to prior benchmarks that rely on single-axis proxies, our approach disentangles surface linguistic complexity from compositional load by independently varying the number of instances and the combination of constraints. This design enables fine-grained, interpretable analysis of cross-modal grounding under increasing compositional complexity, while isolating failure modes without confounding error propagation. As such, it complements existing benchmarks by providing a diagnostic perspective centered on compositional structure.

\section{Proposed method: Imag-Eval} \label{sec:proposed_method}

\subsection{Skills definition}

\subsubsection{Common evaluation skills}

\textbf{(1) Counting} measures how accurately the model generates the requested number of object instances. For example, if the prompt specifies five apples, we check whether exactly five are present. Each object is scored as 0 (failure) or 1 (success), and the final score is the ratio of successful cases to the total number of objects. This is the only skill evaluated at the object level rather than the instance level. An example : \texttt{\{"object": "apple", "count": 5\}}.

\textbf{(2) Spatial Relationships} assesses whether the model correctly positions objects relative to one another (e.g., "apple under table"). The score reflects the average success rate across all instances. An example rule: \texttt{["apple", "under", "table"]}.

\textbf{(3) Color} checks adherence to assigned colors for each object instance. When a color is specified for an object, all its instances must comply. It is embedded within the Counting rule, e.g., \texttt{\{"object": "apple", "count": 5, "color": "blue"\}}.

\textbf{(4) Size} verifies compliance with relative size relationships between instances. An example rule: \texttt{["apple", "smaller", "table"]}.

\textbf{(5) Emotion} evaluates how well the model conveys specified emotions for characters. Success is determined per instance, based on whether the generated emotion matches the prompt. An example rule: \texttt{"a person next to the fire hydrant is \underline{rejoicing}"}.

\textbf{(6) Text} measures the accuracy of generated text against the prompt. Success is evaluated based on how closely the generated text matches the required text. An example : \texttt{A small sign on the wall reads: "Toilet, hot dog: not for sharing"}.

\subsubsection{Under-explored evaluation skill}

\begin{figure}[h]
    \centering
    \includegraphics[width=0.5\linewidth]{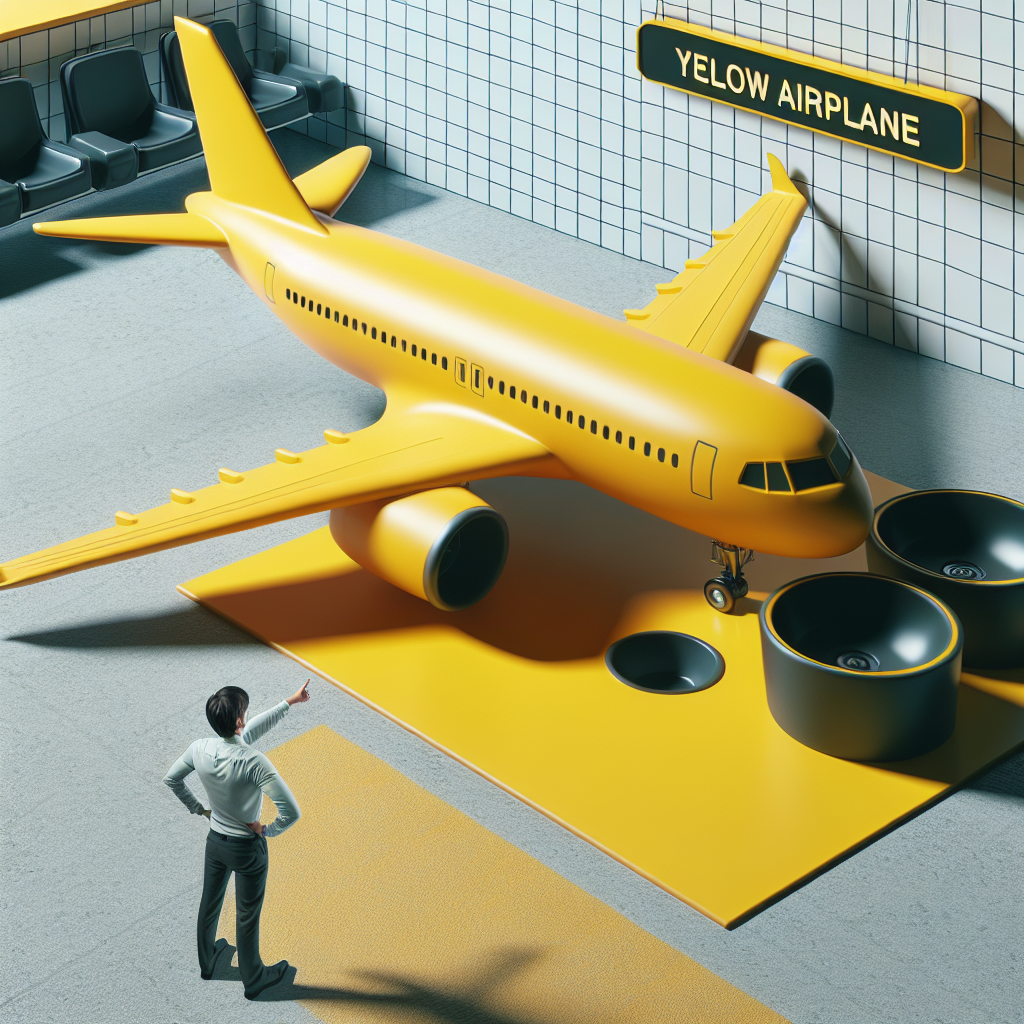}
    \caption{\emph{Cohesiveness}. Dall-E positioned the elements as required, respected spatial relationships, generated the text with the required typo, but created a human instance with an extra arm.}
    \label{fig:cohesiveness_importance}
\end{figure}

\textbf{(7) Cohesiveness.} Determines whether all generated instances are complete and coherent. This binary skill is scored as False if any object lacks essential parts (e.g., a headless human) unless explicitly requested, and true otherwise. As illustrated in Fig.\ref{fig:cohesiveness_importance}, an image can satisfy multiple criteria, but remain unusable in most contexts due to an absence of global \emph{Cohesiveness}. 

We define a lack of \emph{Cohesiveness} as the occurrence of anatomical or structural inconsistencies. These include unprompted deformed, missing, or extra body parts, incomplete objects or surfaces, and physically implausible arrangements (e.g., floating objects).

\subsection{Compositional load}

Our structure involving multiple skill combinations enables controlled evaluation at multiple perspectives: from simpler settings requiring only a small number of skills (two-skill combinations) to more demanding configurations that combine several skills (up to 6-skill combination), with or without robustness testing. We assign up to 5 instances in \textbf{hard} examples, up to 3 instances for \textbf{medium}, and up to 2 in \textbf{easy} examples. We define \emph{compositional load} as the number of grounded constraints 
and their binding to instances. Operationally, we compute the compositional load as the sum of the number of instantiated objects and the number of constraints that must be simultaneously satisfied. Formally, for a prompt $p$, the compositional load $CL(p)$ is defined as:

\begin{equation} 
    \begin{split}
    CL(p) = I(p) + E(p) + T(p) + \\ C_w(p) + Si_w(p)+ Sp_w(p)
    \end{split}
\end{equation} 

where \(I(p)\) is the number of object instances, \(E(p)\) the number of emotion constraints, \(T(p)\) the number of text constraints, and \(C_w(p)\), \(Si_w(p)\), and \(Sp_w(p)\) the instance-weighted counts of color, size, and spatial constraints.

\subsection{Dataset}

\subsubsection{Prompt generation steps} \label{sub:prompt_generation}

\textbf{(1) Meta-prompt initialization.} We begin by constructing a meta-prompt, prefixed with the instruction: \textit{Create a natural language text description for an image that contains the following elements}. This meta-prompt serves as the foundation for generating the final prompt, named \textit{synthetic prompt}.

\textbf{(2) Rule composition.} For each skill under evaluation, we append the relevant parameters to the meta-prompt. For example, when testing \emph{Counting+Color}, we include specifications such as \texttt{\{"object": name, "count": object\_count, "color": assigned\_color\}}, with object names randomly drawn from the COCO dataset \cite{lin_microsoft_2014}. For the \textit{Emotion} skill, emotions are assigned exclusively to human figures within the scene, ensuring each emotion is tied to a single individual. For \textit{Size} and \textit{Spatial Relationships}, we append relational instructions (e.g., \texttt{\{A, smaller, B\}} or \texttt{\{A, under, B\}}) after the object declarations to maintain logical consistency. For the \textit{Text} skill, we include a textual element (e.g., a sign) that references specific objects from the prompt.

\textbf{(3) Object assignment.} We then assign object counts based on difficulty level: up to 5 objects per skill (3 for emotion) in hard examples, up to 3 objects (2 for emotion) in medium examples, and up to 2 objects (1 for emotion) in easy examples. This assignment is streamlined by our JSON-based skill architecture, which we convert into a string and append to the meta-prompt.

\textbf{(4) Semi-modular prompt generation.} Finally, we use a text generation model to produce a description of the hypothetical image. To increase robustness in challenging cases, we create a version of each test with lexical and semantic perturbations, by introducing typos and synonyms. For each sample, the meta-prompt is dynamically constructed based on the skills being evaluated. With this approach, we generated a dataset of 1,140 prompts, manually verified with respect to the JSON rules. Dataset statistics are available in App.\ref{app:dataset_statistics}.

Fig \ref{fig:prompt_example} illustrates an example of a synthetic prompt generated from such a meta-prompt, specifically for a test combining Color, Emotion, and Text skills at an easy level. This example highlights how elements from the meta-prompt are naturally integrated into the final text description. To generate these text descriptions, we selected models from the literature that demonstrated strong comprehension abilities, specifically those achieving over 80\% accuracy on the MMLU-Pro benchmark \cite{hendrycks_measuring_2021}. This benchmark's focus on understanding capabilities made it particularly relevant for our needs, as it directly relates to a model's ability to understand the context. We also considered resource constraints and model availability during our selection process. Focusing on recent models (2024 and later), we generated 30 sample synthetic prompts using local Qwen3-30B-A3B-Instruct-2507 \cite{yang_qwen3_2025}, Mindlink-32B, Deepseek R1 \cite{guo_deepseek-r1_2025}, and GPT-5 \cite{singh_openai_2025}. After comparing the outputs, we selected GPT-5, as its resulting prompts demonstrated fewer errors and greater originality. \textbf{However, our JSON rule-based structure allows for generating with any other model. Minor tweaks can also be applied to the code to increase the number of levels, instances, or rules, by modifying a parameter within the scripts. A description on the "how" is available in App.\ref{app:modifying_levels}.}

It is important to note that proprietary LLM APIs and their parameters may evolve over time, which can introduce minor variations in prompt generation. Nevertheless, our experiments indicate that the relative differences in compositional load are generally preserved across such variations.

\begin{figure}[htbp]
    \centering
    \setlength{\fboxsep}{3pt} 
    \fbox{%
        \parbox{0.9\linewidth}{%
    \texttt{A street scene with 
    \textcolor[HTML]{E69F00}{a bright yellow parking meter} 
    in the foreground and a large
    \textcolor[HTML]{CC79A7}{pink bear} 
    standing on the sidewalk; 
    \textcolor[HTML]{D55E00}{a surprised person} 
    nearby with mouth agape and hands raised is clearly reacting to the bear; 
    \textcolor[HTML]{0072B2}{a sign on a lamppost prominently reads "Pink Bear Ahead".}}
            }%
        }
    \caption{Prompt example : \textit{Color+Text+Emotion} skills. Highlighted parts are derived from our JSON rules.}
    \label{fig:prompt_example}
\end{figure}

\subsubsection{Image generation} \label{sub:image_generation}

\begin{figure}[ht]
    \centering
    \begin{subfigure}{0.48\columnwidth}
        \centering
        \includegraphics[width=0.8\linewidth]{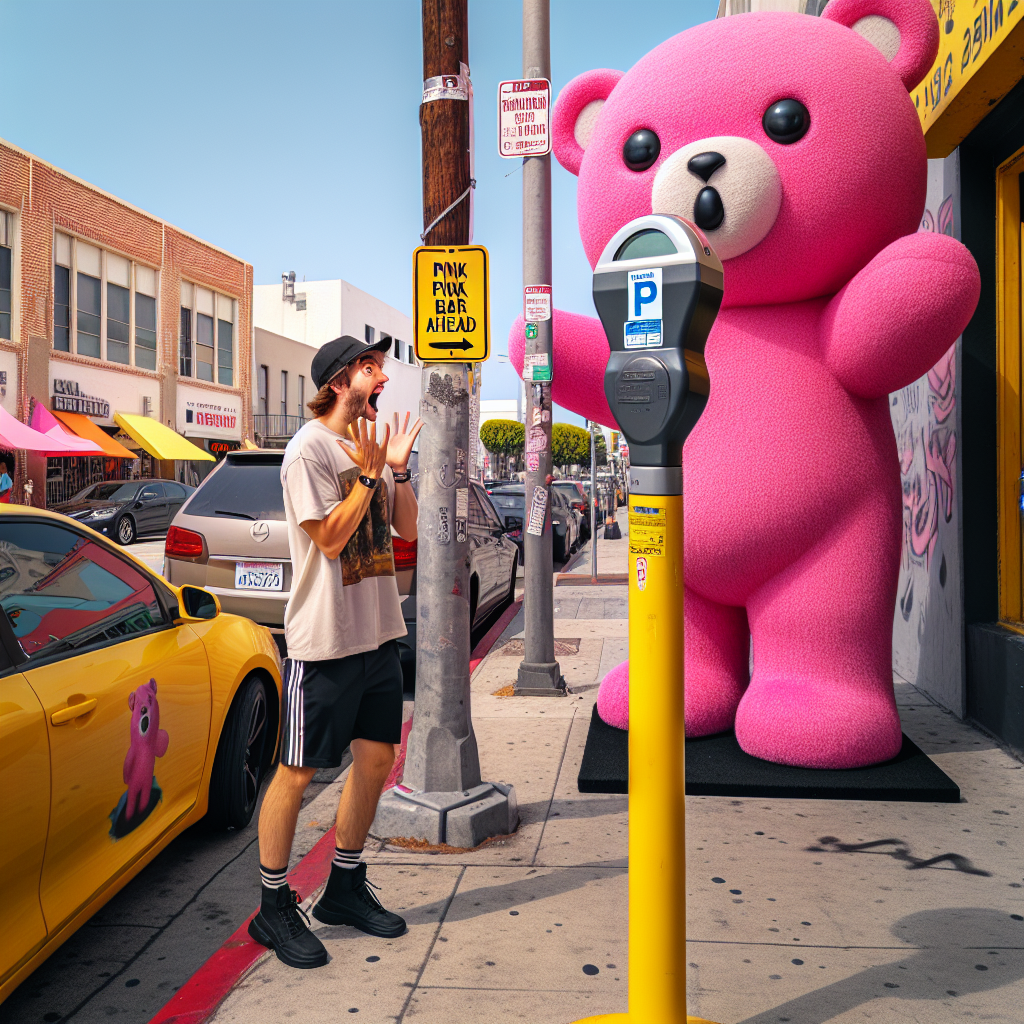}
        \caption{Dall-E 3.}
        \label{fig:dalle_example}
    \end{subfigure}
    \hfill
    \begin{subfigure}{0.48\columnwidth}
        \centering
        \includegraphics[width=0.8\linewidth]{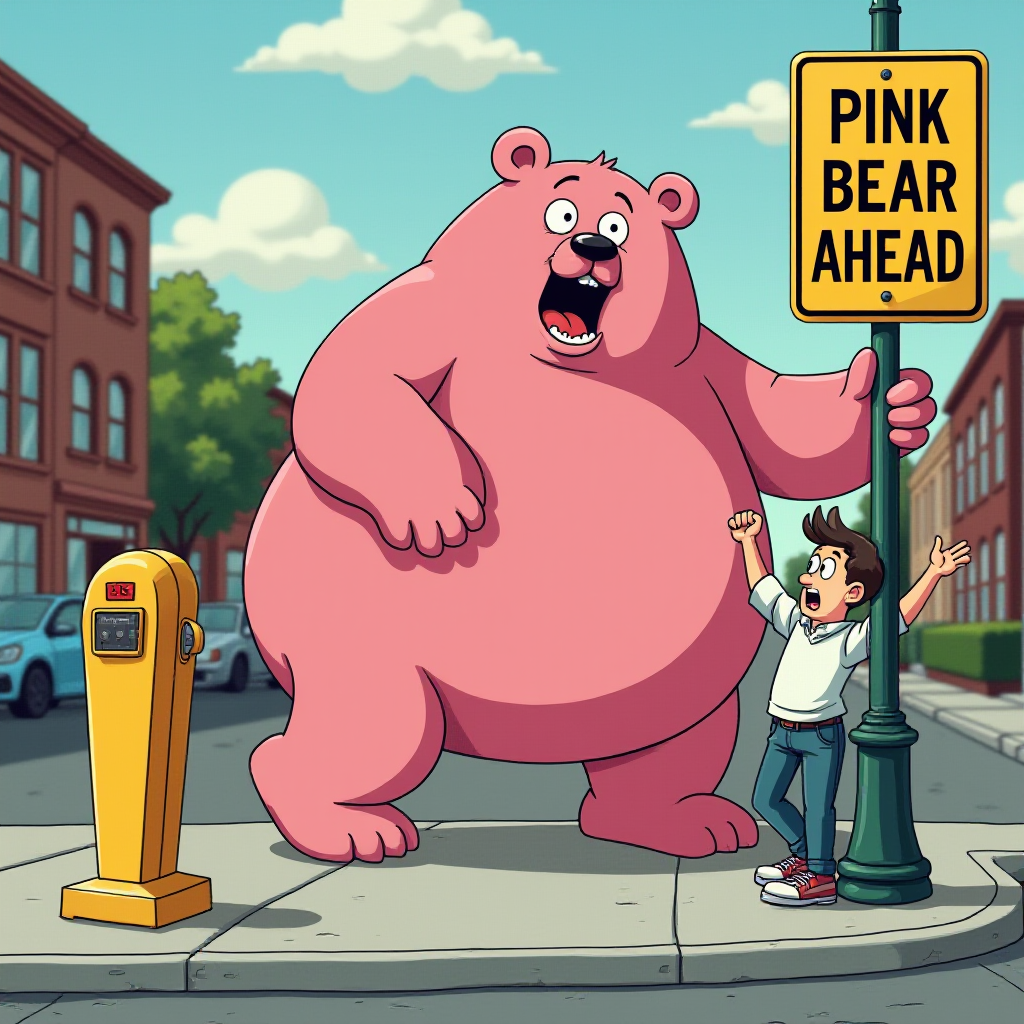}
        \caption{FLUX.}
        \label{fig:flux_example}
    \end{subfigure}

    \par\medskip

    \begin{subfigure}{0.48\columnwidth}
        \centering
        \includegraphics[width=0.8\linewidth]{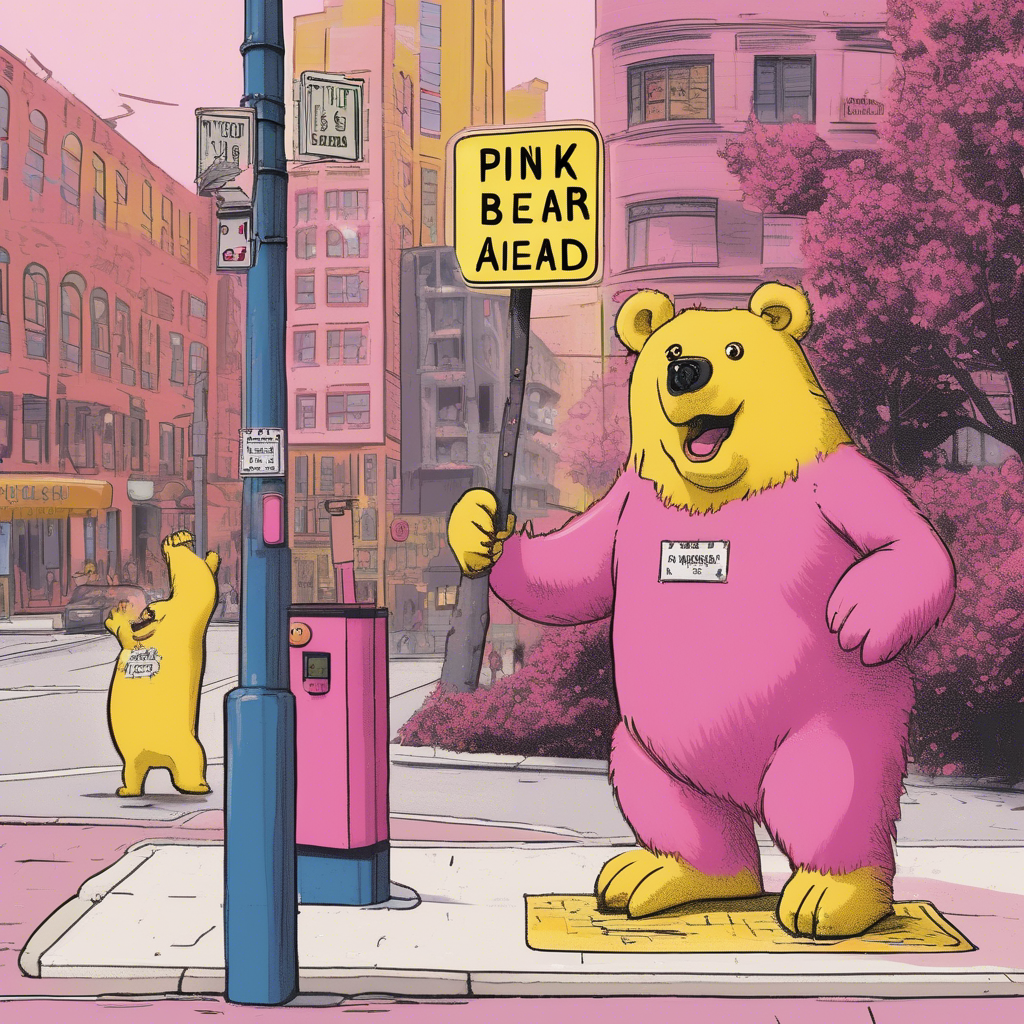}
        \caption{Stable Diffusion.}
        \label{fig:stable_diffusion_example}
    \end{subfigure}
    \hfill
    \begin{subfigure}{0.48\columnwidth}
        \centering
        \includegraphics[width=0.8\linewidth]{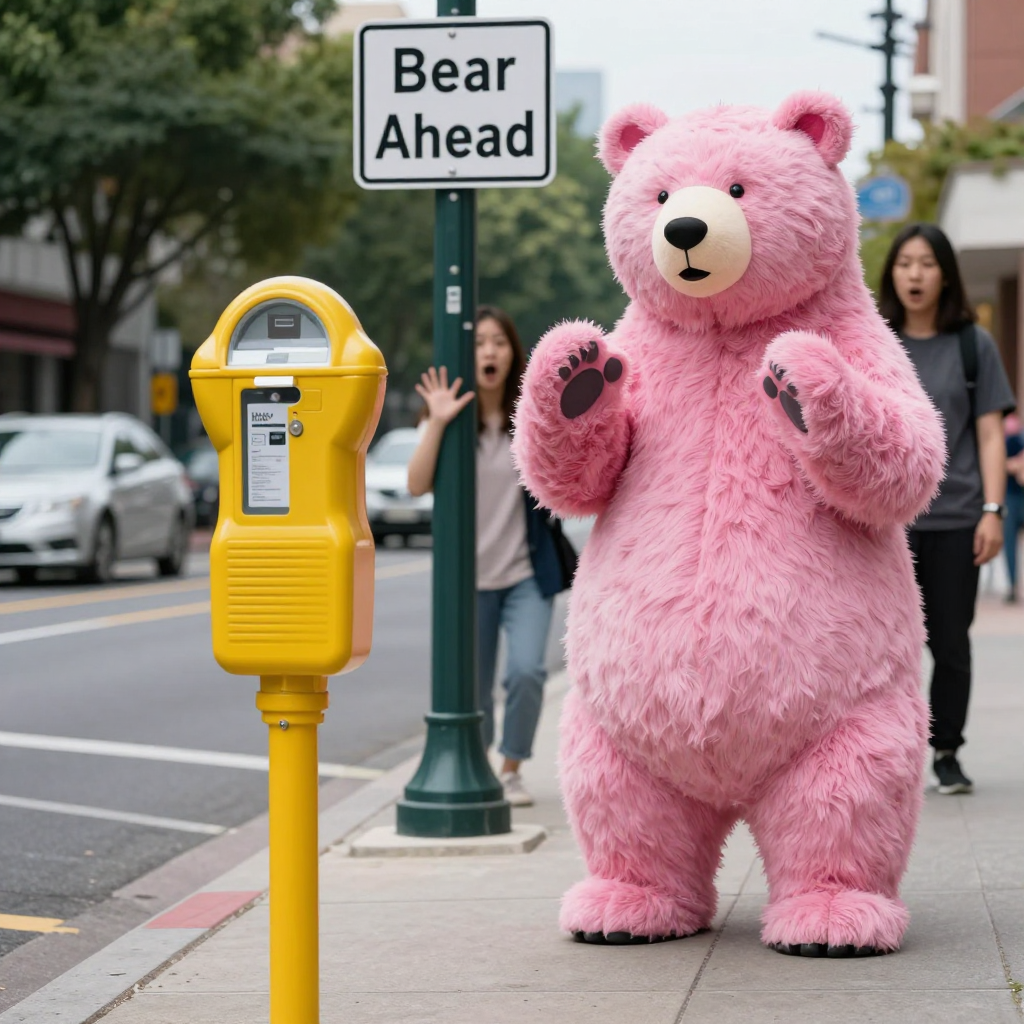}
        \caption{Z-Image Turbo.}
        \label{fig:z_image_example}
    \end{subfigure}

    \caption{Examples generated by various models.}
    \label{fig:images_examples}
\end{figure}

For image generation, we evaluated several models from the literature, considering factors such as parameter count, VRAM consumption, the balance between open-source and proprietary models, and publication year. Given the rapid advancements in image generation over the past three years, we excluded models whose latest versions were released before 2022. We prioritized models supported by whitepapers or peer-reviewed studies, as well as those developed by reputable sources. Fig.\ref{fig:images_examples} shows a generated sample from the synthetic prompt for reference, for different models. All generations were conducted on a shared compute infrastructure comprising one NVIDIA RTX 6000 Ada Generation GPU with 49\,GB of VRAM and two NVIDIA RTX 5000 GPUs with 32\,GB of VRAM each. The system possessed an Intel Xeon w5-3435X (4.70\,Ghz) CPU featuring 32 threads.

Based on our resource constraints for model loading and inference, for proprietary models, we selected DALL-E 3 \cite{betker2023dalle3} from OpenAI, partly to assess if a model from the same developer as our prompt generation model would perform better, and Gemini 3.1-Flash-preview \cite{gemini31flashimage2026} for which we acquired a license. For open-source alternatives, we included Z-Image Turbo \cite{cai_z-image_2025}, Stable Diffusion XL \cite{podell_sdxl_2024}, FLUX 1.0-dev \cite{blackforestlabs2024fluxdev} and Stable Cascade based on the Würstchen architecture \cite{pernias_wurstchen_2024}, to ensure a diverse set of models. Although we originally intended to use Qwen-Image \cite{wu_qwen-image_2025} and Hunyuan-Image-3.0 \cite{cao_hunyuanimage_2025} from Tencent, they were ultimately excluded as we were unable to load their full versions and wished to avoid comparisons with unofficial or distilled variants. To minimize developer bias, since we already included two models from the same distribution, we excluded GLM-Image which shares developers with Z-Image Turbo (Alibaba), and DeepFloyd \cite{saharia_photorealistic_2022} (Stability AI, Stable Cascade).

We used the optimal inference parameters from the documentation of each model, as detailed in App.\ref{app:image_generation}.  Following this approach, we generated >5,500 images for the 6 models. When a prompt exceeded the maximum sequence length of a model, we used SD-Embed \cite{sd_embed_2024} (Apache License 2.0) to encode the text as learned embeddings.   

\subsubsection{Annotation process} \label{sub:annotation_process}

To ensure diverse perspectives in the annotation process, we engaged 14 annotators from varied cultural and educational backgrounds (App.\ref{app:instructions_and_recruitment} for more details). Each distribution was annotated by three individuals, and annotations were merged upon manual verification, except for the less-objective \emph{Cohesiveness} and \emph{Emotion}, for which we solve disagreements based on majority votes. Our first merging method focused on annotator agreement, however, we found that some specific instances were too much confusing for annotators, such as those represented in App.\ref{app:confusing_cases} where humans are mixed with cats, which led to  a global confusion. We proceeded to manual validation to ensure fair results. 

Files were named according to the convention: \texttt{[model\_name]\_[skills]\_[level]\_[prompt number]\_[robust/nothing]}. For example, the image in Fig.\ref{fig:z_image_example} is named \textit{z\_image\_turbo\_2\_easy\_001\_0.png}, indicating an easy-level prompt generated by Z-Image-Turbo for the \textit{Color + Emotion + Text} skill combination. For the annotation process, each skill combination was assigned a specific code (e.g., 12 for "color+size"). Annotators first identified the model name to locate the corresponding folder (e.g., \textit{flux}) and then applied the criteria outlined in Sec.\ref{sec:proposed_method}, along with column-specific rules. Annotation rules varied by skill. For \textit{Counting}, values were recorded in the format \texttt{Instances\_generated,Instances\_required} for each object, separated by semicolons, allowing us to track over- or under-generation. For \textit{Text}, annotators transcribed all visible text from signs or boards, separated by semicolons, with each sign’s text on a new line. This approach helped capture instances where the model generated additional or incorrect text. For other skills, annotators recorded success rates, except for \textit{Cohesiveness}, which was strictly binary (TRUE or FALSE). This process resulted in \textbf{+8,000 validated annotated raw rules}.

\section{Experiments results} \label{sec:experiments_results}

\subsection{Metrics} \label{sub:metrics}

In addition to the success rates presented in Sec.\ref{sec:proposed_method}, we utilize the Word Error Rate (WER) \cite{morris2004and} to measure the discrepancy between the generated text and the target text specified in the prompt. A lower WER indicates higher accuracy, with a value of 0 representing an exact match between the generated and reference texts. We opted for WER over embedding-based metrics such as cosine similarity because our objective was to assess the exact lexical match between the generated and target texts, rather than their semantic similarity.

\subsection{Evaluation settings}

As our benchmark comprises approximately 114 possible skill combinations, we focus on the \textit{All Skills} setting in the remainder of this paper. This configuration represents the highest compositional load, requiring models to satisfy all evaluation dimensions simultaneously. Unless otherwise specified, the reported results are based on manual annotations and jointly assess Counting, Spatial Relations, Size Relations, Color Attribution, Emotion Attribution, Text Rendering, and Cohesiveness.

\subsection{Overall results} \label{sub:overall_results}

The results presented in Tab.\ref{tab:overall_results} indicate that the Gemini-Flash-3.1 model achieved superior overall performance across most metrics (4/7) when considering all difficulty levels, with the exception of the Counting skill. However, a closer examination of \emph{Cohesiveness} reveals that it failed to generate coherent images in 32\% of the cases. WER values greater than 1 suggest that most models tended to generate either significantly longer sequences or additional sentences compared to the required text.

\begin{table*}[htbp]
  \centering
  \begin{tabular}{m{0.16\textwidth}m{0.11\textwidth}m{0.11\textwidth}m{0.11\textwidth}m{0.11\textwidth}m{0.11\textwidth}m{0.11\textwidth}}
    \toprule
    \textbf{Skill} & Z-Image-Turbo & FLUX 1.0 & Dall-E 3 & \textbf{Gemini-Flash} & SDXL & SC \\
    \midrule
    $\uparrow Counting$ & 0.55 & 0.56 & 0.42 & \textbf{0.70} & 0.17 & 0.13 \\
    \midrule
     $\uparrow Spatial$ & \textbf{0.79} & 0.69 & 0.59 & 0.73  & 0.34 & <0.01 \\
    \midrule
     $\uparrow Size$ & 0.84 & 0.62 & 0.70 & \textbf{0.85} & 0.15 & 0.20 \\
     \midrule
     $\uparrow Emotion$ & $0.80(0.94)$ & $0.67(0.92)$ & $0.28(1.00)$ & $\textbf{0.98}(0.98)$ & $0.09(1.00)$ & $0.10(1.00)$ \\
     \midrule
     $\uparrow Color$ & \textbf{0.98} & 0.88 & 0.69 & 0.96 & 0.16 & 0.36 \\
     \midrule
     $\uparrow Cohes.$ & $0.50(0.90)$ & $0.76$(0.90) & $\textbf{0.99}(0.99)$ & $0.68(0.98)$ & $0.26(1.0)$ & $0.35(1.0)$ \\
     \midrule
     $\downarrow Text (WER)$ & 0.49 & 2.46 & 4.25 & \textbf{0.29} & 5.55 & 4.31 \\
  \bottomrule
  \end{tabular}
  \caption{Results for the "all skills" test, on manual annotation. FLUX 1.0=FLUX 1.0-dev, Gemini-Flash=Gemini-Flash-3.1-preview, SDXL=Stable Diffusion XL, SC=Stable Cascade. For the interpretative skills Emotion and Cohesiveness, values in parentheses report pairwise Cohen's \(\kappa\), averaged across annotator pairs. Detailed results are available in App.\ref{app:overall_results}. For most categories, a higher score means a better evaluation, except for WER.}
  \label{tab:overall_results}
\end{table*}

\subsection{Analysis by level and number of skills} \label{sub:granular_level_and_combined_complexity}

\begin{figure*}[htb]
    \centering

    \begin{subfigure}{0.4\textwidth}
        \centering
        \includegraphics[width=\linewidth]{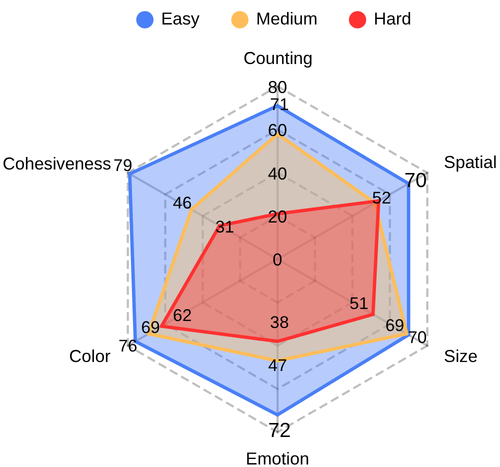}
        \caption{Skills by level.}
        \label{fig:results_skills_by_level}
    \end{subfigure}
    \hfill
    \begin{subfigure}{0.4\textwidth}
        \centering
        \includegraphics[width=0.7\linewidth, height=0.2\textheight]{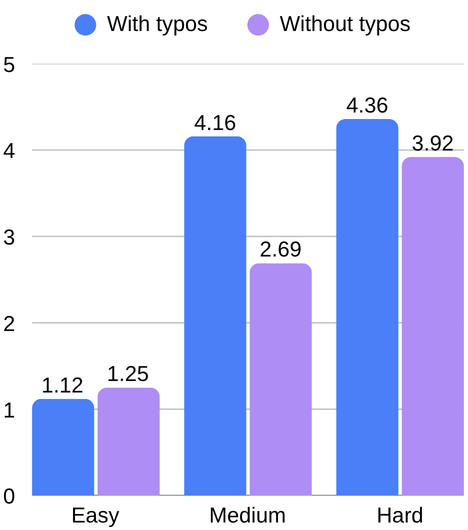}
        \caption{WER with and without typos.}
        \label{fig:impact_typos}
    \end{subfigure}

    \begin{subfigure}{0.4\textwidth}
        \centering
        \includegraphics[width=\linewidth]{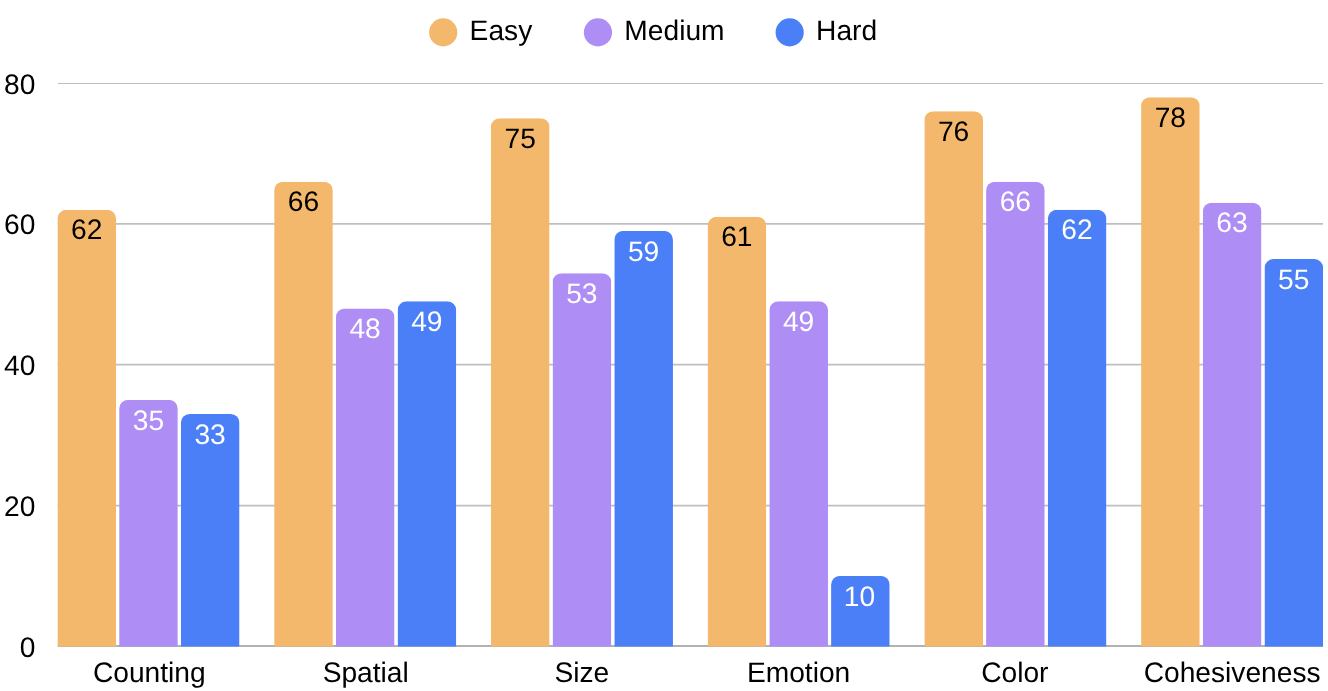}
        \caption{Results on similar prompt lengths.}
        \label{fig:results_prompt_length}
    \end{subfigure}
    \hfill
    \begin{subfigure}{0.45\textwidth}
        \centering
        \includegraphics[scale=0.15]{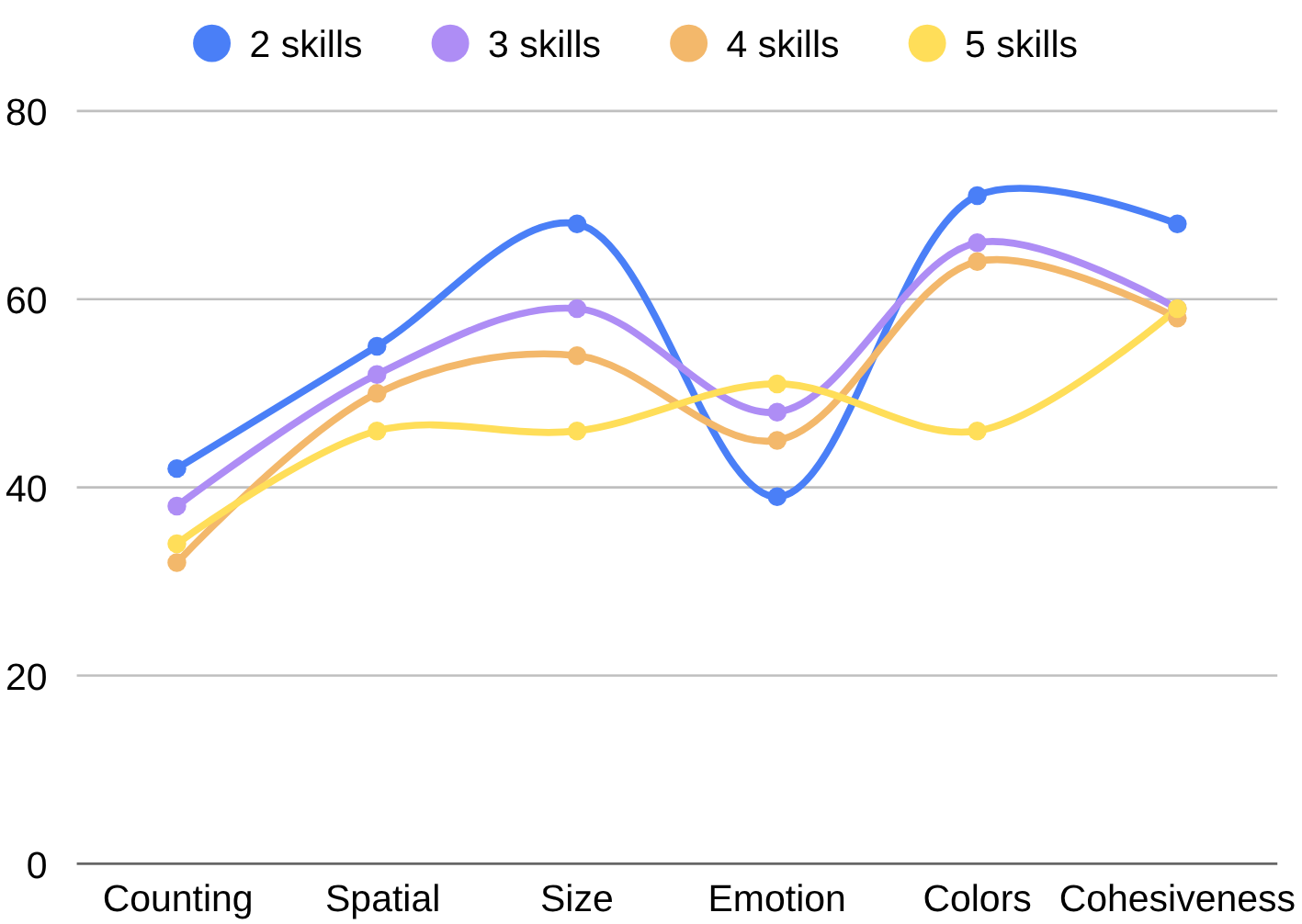}
        \caption{Analysis by number of skill combinations (rules).}
        \label{fig:results_skills_by_combination_level}
    \end{subfigure}

    \caption{Breakdown of evaluation metrics (avg) on multiple axis. More complete results available in App.~\ref{app:overall_results_number_of_skills}.}
    \label{fig:results_analysis_by_level}
\end{figure*}

Analyzing performance across difficulty levels further elucidates the conditions under which models tend to fail, particularly as the required number of generated instances increases. Fig.\ref{fig:results_skills_by_level}, which aggregates results across all evaluated models, shows a highly non-uniform degradation across skills. The most pronounced decline from \emph{easy} to \emph{hard} occurs for the \emph{Text} skill, with a 243\% increase in error rate, indicating a substantial rise in hallucinated textual content. This is followed by \emph{Counting}, \emph{Cohesiveness}, and \emph{Emotion}, which exhibit accuracy decreases of 71\%, 60\%, and 47\%, respectively. In contrast, \emph{Color} remains the most robust skill, with a comparatively modest drop of 19\%. 

Further insight is obtained by analyzing skill performance as a function of the number of combined skills. As shown in Fig.\ref{fig:results_skills_by_combination_level}, even at a fixed difficulty level (here, \emph{easy} level), accuracy typically decreases as the number of required skill combinations increases. This pattern reinforces our core hypothesis that generation quality is jointly impacted by both the number of instances to generate and the compositional load induced by multiple simultaneous constraints. While Fig.\ref{fig:results_skills_by_combination_level} focuses on a single  difficulty level, the same degradation is observed across all difficulty levels.

\subsection{Prompt length vs Compositional load} \label{sub:prompt_length_vs_rule_complexity}

While the dataset statistics (App.~\ref{app:dataset_statistics}) indicate that higher difficulty levels are generally associated with longer prompts and increased linguistic complexity (e.g., higher adjective density and greater use of prepositions), these aggregate trends remain incomplete. To control for potential confounding effects of prompt length, we isolate a subset of prompts with comparable lengths across all difficulty levels. Specifically, we estimate the probability distributions of prompt lengths for each level, identify the region with maximal overlap across distributions, and select the interval corresponding to the highest shared density. This procedure yields a length range of 50--77 tokens, comprising approximately 100 prompts per level. Fig.\ref{fig:results_prompt_length} reports results restricted to this interval, aggregated across all models, on >2K annotated rules. The same trend persists: performance across most skills consistently degrades as difficulty increases. Importantly, we show in Tab.~\ref{tab:textual_stats_similar} that within this controlled subset, standard measures of linguistic complexity do not vary significantly with difficulty level.

Moreover, the standardized multivariate regression analysis shown in Tab. ~\ref{tab:regression_results_multivariate} indicates that compositional load accounts for a substantially larger share of performance variation than prompt length. Across all structured skills, compositional load yields larger standardized coefficients and lower p-values, while its confidence intervals remain consistently below zero, providing strong evidence of a negative association between increasing compositional complexity and model accuracy. By comparison, the effect of prompt length is generally smaller and less stable. Furthermore, VIF values below 5 for all predictors suggest that multicollinearity is limited, supporting the interpretation that compositional load contributes explanatory power beyond that captured by prompt length alone.

\begin{table}[htb]
    \centering
        \begin{tabular}{cccc}
            \toprule
            \textbf{Criteria (avg)} & \textbf{Easy} & \textbf{Medium} & \textbf{Hard} \\
            \midrule
            Length & 65.37 & \underline{63.15} & \textbf{67.31} \\
            \midrule
            Entities & 18.37 & \underline{17.02} & \textbf{18.94}  \\
            \midrule
            Relations & 4.78 & \underline{4.66} & \textbf{4.92} \\
            \midrule
            Parse depth & 2.37 & \textbf{2.47}& \underline{2.25} \\
            \midrule
            Prepositions & 6.85 & 7.04  & \textbf{7.22} \\
            \bottomrule
        \end{tabular} 
    \caption{Statistics on prompts of similar lengths, obtained using NLTK tokenizer. Entities=Nouns, Relations=Number of connectors (with, on, under...), Parse depth=Number of clauses (which, that...) and sentences.}
    \label{tab:textual_stats_similar}
\end{table}

\begin{table*}[h]
    \centering
    \begin{tabular}{ccccc}
        \toprule
        \textbf{Skill} & \textbf{Coefficients} & \textbf{P-values} & \textbf{Confidence intervals} & \textbf{VIF (L,CL)} \\
        \midrule
         \emph{Counting} & \makecell{(L) $+0.0216$ \\ \textbf{(CL)} $\bm{-0.3634}$} & \makecell{(L) $0.61$ \\ \textbf{(CL)} $\bm{9.8\times10^{-17}}$} & \makecell{(L) $[-0.06;0.10]$ \\ (CL) $[-0.44;-0.27]$} & 3.52 \\
         \midrule
         \emph{Spatial} & \makecell{(L) $+0.0341 $ \\ \textbf{(CL)} $\bm{-0.1910}$} & \makecell{(L) $0.29$ \\ \textbf{(CL)} $\bm{4.89\times10^{-9}}$} & \makecell{(L) $[-0.029;0.09]$ \\ (CL) $[-0.25;-0.12]$} & 2.83 \\
         \midrule
         \emph{Size} & \makecell{(L) $+0.0418$ \\ \textbf{(CL)} $\bm{-0.1883}$} & \makecell{(L) $0.24$ \\ \textbf{(CL)} $\bm{2.13\times10^{-7}}$} & \makecell{(L) $[-0.02;0.11]$ \\ (CL) $[-0.25;-0.11]$} & 3.15 \\
         \midrule
         \emph{Emotion} & \makecell{\textbf{(L) $\bm{-0.1203}$} \\ (CL) $-0.0130$} & \makecell{\textbf{(L) $\bm{0.001}$} \\ (CL) $0.73$} & \makecell{(L) $[-0.19;-0.04]$ \\ (CL) $[-0.08;0.06]$} & 3.03 \\
         \midrule
         \emph{Colors} & \makecell{(L) $-0.0074$ \\ \textbf{(CL)} $\bm{-0.1296}$} & \makecell{(L) $0.82$ \\ \textbf{(CL)} $\bm{0.0001}$} & \makecell{(L) $[-0.07;0.05]$ \\ (CL) $[-0.19;-0.06]$} & 3.50 \\
         \midrule
         \emph{Text (WER)} & \makecell{(L) $+0.0910$ \\ \textbf{(CL)} $\bm{+0.1548}$} & \makecell{(L) $0.017$ \\ \textbf{(CL)} $\bm{0.000059}$} & \makecell{(L) $[0.01;0.16]$ \\ (CL) $[0.07;0.23]$} & 3.07 \\
         \midrule
         \emph{Cohesiveness} & \makecell{\textbf{(L) $\bm{+0.0732}$} \\ (CL) $+0.0159$} & \makecell{\textbf{(L) $\bm{0.008}$} \\ (CL) $0.56$} & \makecell{(L) $[0.01;0.12]$ \\ (CL) $[-0.03;0.07]$} & 2.88 \\
         \bottomrule
    \end{tabular}
    \caption{Results of the multivariate regression and correlation analyses: skill accuracy vs (prompt length and compositional load). Abbreviations: CL=Compositional Load, L=Length, VIF = Variance Inflation Factor. All the coefficients were standardized prior to model fitting. Across most structured skills (Counting, Spatial, Size, Color, and Text), compositional load has a strong and statistically significant effect, whereas prompt length exhibits a weaker effect. In contrast, for interpretative skills (Emotion and Cohesiveness), this relationship is less pronounced.}
    \label{tab:regression_results_multivariate}
\end{table*}

\subsection{Analysis of TIIF-Bench}

TIIF-Bench \cite{wei_tiif-bench_2025} reports a correlation between prompt length and the average skill accuracy. However, a closer examination of their prompt construction reveals an intriguing pattern.

Consider the example in Fig.~\ref{fig:tiif_example}. By analogy with our prompt construction, the shorter prompt can be interpreted as involving two instances (\emph{pig},\emph{cup}) and a single \emph{Spatial} rule (\emph{to the left of}). In contrast, the longer variant includes the original compositional load and increases it with additional descriptive details such as texture attributes. In our benchmark, while such descriptive elements may be included, they are not systematically scaled with difficulty. Instead, variations in prompt length arise mainly from the controlled factors. Our initial analysis compared the lexical content of the long and short prompt variants and showed that the additional content was predominantly related to scene atmosphere, texture descriptions, and contextual details. We subsequently formalized these recurring additions as additional "rules" and re-analyzed the benchmark from that perspective. Under this procedure, more than 90\% of the analyzed prompt pairs exhibited an increase in compositional content beyond simple length expansion.

Although prompt length can influence model performance by increasing the burden on the text encoder to faithfully capture constraints, we argue that disentangling length from compositional structure is critical for a more interpretable assessment.

\begin{figure}[H]
    \centering
    \setlength{\fboxsep}{3pt} 
    \fbox{%
        \parbox{0.9\linewidth}{%
        \texttt{\textbf{(Short):} \textcolor[HTML]{E69F00}{A cup} is positioned \textcolor[HTML]{CC79A7} {to the left of} \textcolor[HTML]{CC79A7} {a pig}. \\
        \textbf{(Long):} Positioned \textcolor[HTML]{CC79A7}{tranquilly to the left} of \textcolor[HTML]{D55E00}{the pig}, which stands as \textcolor[HTML]{E69F00}{a silent and innocent observer}, the \textcolor[HTML]{CC79A7}{unassuming cup}, \textcolor[HTML]{0072B2}{a simple vessel of ceramic or maybe porcelain}, rests quietly, its presence understated \textcolor[HTML]{D55E00}{yet somehow integral to the quietude of the scene, where each object seems to hold its breath} in the gentle stillness that pervades the atmosphere.}}
    }%
    \caption{Example of two versions of a prompt, from TIIF-Bench\cite{wei_tiif-bench_2025}. Parts of the text that are highlighted represent potential "rules" or instances.}
    \label{fig:tiif_example}
\end{figure}

\subsection{T2I models struggle at typos} \label{sub:robust_vs_nonrobust}

Fig.\ref{fig:impact_typos} compares text rendering performance for prompts with and without injected typos. Across most difficulty levels, they exhibit substantially larger deviations in generated text. A closer inspection of failure cases under typo-conditioned prompting shows that, beyond hallucinations, models often either introduce alternative spelling errors rather than the specified ones (30\% of cases). 

\section{Discussion} \label{sec:discussion}

\subsection{The importance of human annotations} \label{sub:automation}

We developed an automated pipeline to streamline the annotation process. \textbf{Object detection, monocular depth estimation and and instance segmentation} using YOLO-26 \cite{jocher2026ultralyticsyolo26unifiedrealtime} (AGPL-3.0), extracts bounding boxes and class labels, depth maps and detection masks, enables automated evaluation of the \emph{Counting, Spatial} and \emph{Size} skills. \emph{Color, Emotion} and \emph{Cohesiveness} rely on a Qwen3-VL \cite{yang_qwen3_2025} inference (Apache License 2.0).

However, the pipeline struggled significantly with \emph{Cohesiveness}, for which (VQA) model alignment with human judgment rarely exceeded 70\%. We explored various methods to improve those assessments, drawing on recent MLM-as-judge studies \cite{chen_mllm-as--judge_2024}, including instance-level evaluation and prompt reformulation. However, results remained unsatisfactory. Traditional neural networks based frameworks such as DeepFace \cite{serengil_lightface_2020} also failed to deliver significant improvements. This further reinforces the importance of human oversight for interpretative dimensions.

\subsection{The state of current evaluation}

Our findings, especially when reviewing the state-of-the-art, raised an intriguing question: \textbf{\emph{Are our current evaluation methods too "static"?}} To the best of our knowledge, most existing benchmarks (including ours) rely on a fixed set of skills or predefined combinations thereof. A promising direction for future evaluation lies in the development of \emph{modular} benchmarks, in which skills can be added or removed on demand and corresponding evaluation prompts can be generated accordingly. Such adaptability would enable continuous refinement of evaluation protocols as model capabilities evolve.

\section{Conclusion}

This work introduced \textsc{Imag-Eval}, a controlled and interpretable evaluation framework aimed at diagnosing the instruction-following capabilities of Text-to-Image (T2I) models. Motivated by the limitations of existing benchmarks, we focused on disentangling surface linguistic complexity from compositional difficulty, and on capturing failure modes that are often overlooked in standard skill-based evaluations, such as global incoherence and incomplete object generation. We further introduced a benchmark dataset comprising 1{,}140 prompts and +8{,}000 combined rules, and validated our approach through extensive experiments involving more than 6{,}000 generated images, 14 annotators, and +8{,}000 annotated rules. These results demonstrate both the feasibility and the diagnostic value of controlled, compositional evaluation protocols.

Our empirical findings highlight that model performance is primarily driven by compositional load (the number of grounded rules and their binding to instances) rather than by surface-level linguistic properties such as prompt length, for structured skills. More broadly, they underscore the limitations of evaluation practices that rely on single-axis proxies for difficulty, and motivate the need for benchmarks that provide a more structured decomposition of multimodal reasoning.

Looking forward, an important direction is the design of flexible and modular benchmarks for multimodal evaluation. While our current framework already enables controlled variations (e.g., adjusting instance counts or regenerating prompts with fixed constraints), a natural extension is to support fully modular skill composition, where evaluation dimensions can be dynamically added, removed, or recombined. Such flexibility would enable more systematic stress-testing of model capabilities and foster the development of more robust and interpretable multimodal systems.

\section*{Acknowledgments}

This work was funded by Talan France through its Research and Innovation Center. We also thank all annotators for their invaluable contributions to data collection and quality assurance, in particular Mariem AMMAR, Dodji Idelphonse DECADJEVI and Samia TEKAL, for their exceptional commitment to the project.

\section*{Limitations}

While our evaluation framework demonstrates promising results, it is currently limited to prompts constructed from COCO object categories and attributes emotions exclusively to human instances. Nevertheless, COCO covers a broad spectrum of common objects, enabling the use of a wide range of off-the-shelf object detectors for automated evaluation, as most contemporary detectors are trained and benchmarked on COCO.

Moreover, all images in this study were generated using a single random seed (42). Evaluating multiple seeds would have required generating and annotating a substantially larger number of samples, resulting in annotation costs beyond the resources available for the present work. Nonetheless, robustness across random initializations is an important consideration, and future versions of the leaderboard will include multi-seed evaluations together with standard deviation estimates.

A further limitation is that our method evaluates single-prompt instructions rather than multi-prompt, incremental instruction sequences. Although some models may perform better with step-by-step guidance, we lack a clear methodology for determining the optimal order of rule presentation and its impact across models, due to the potential number of combinations of orders and skills. 

Finally, our evaluation may be subject to two sources of bias.

\textbf{\emph{(1) Language bias.}} 
All prompts in Imag-Eval are formulated in English. Although all evaluated models officially support English-language inputs, our findings may not directly generalize to languages with substantially different typological, morphological, or syntactic properties. Extending the benchmark to multilingual settings is therefore an important direction for future work.

\textbf{\emph{(2) Model-identity bias.}} 
The annotation protocol did not fully blind annotators to model identity, as generated images were organized using filenames and folders that included model names. This design choice simplified annotation management and downstream analysis, but may have introduced bias in subjective judgments. We partially mitigate this concern through multiple annotators, majority voting for interpretative skills, and manual validation of ambiguous cases. Moreover, most annotators had limited familiarity with the evaluated T2I systems beyond widely known commercial models such as Gemini-Flash-3.1. Nevertheless, future versions of the benchmark should adopt fully anonymized filenames and randomized annotation interfaces.

\bibliography{neurips_ed_2026}

\appendix

\section{Modifying levels and instance count} \label{app:modifying_levels}

\begin{figure*}[h]
    \centering
    \includegraphics[width=0.7\linewidth]{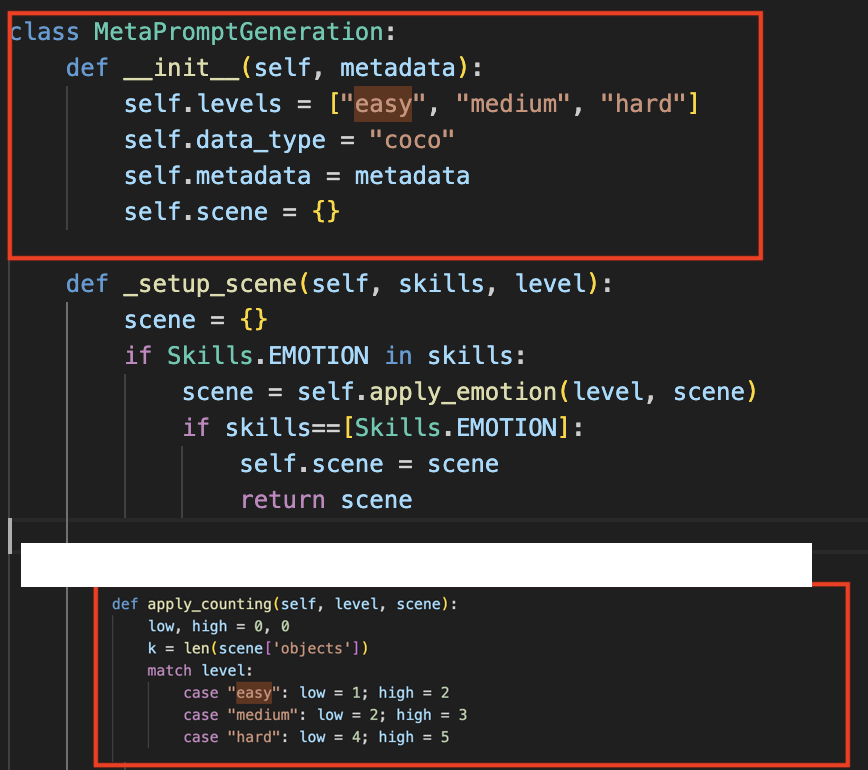}
    \caption{Configuration parameters controlling difficulty granularity and the number of generated instances.}
    \label{fig:modifying_parameters}
\end{figure*}

As discussed earlier, our JSON-based architecture is designed to flexibly redefine both the granularity of difficulty levels and the number of instances to generate at each level. As illustrated in Fig.\ref{fig:modifying_parameters}, adding any level to the list of predefined levels (top panel) and any cases within the cases (bottom panel) allow for re-defining levels. Re-running the main prompt-generation script after such changes automatically produces a new set of prompts reflecting the updated parameters.

In addition, the classes that govern prompt generation within the scripting interface (\texttt{llm\_interfaces}) are model-agnostic. Users may specify their own API keys and model parameters via a \texttt{.env} file, provided the required configuration fields are defined. As a result, the framework is not restricted to GPT-5 or OpenAI models, and can be readily extended to evaluate a wide range of proprietary or open-source text-to-image systems.

\section{Image generation hyperparameters} \label{app:image_generation}

Tab.\ref{tab:app_image_generation} showcases hyperparameters used during our image generation process. All parameters used, except random seeds, were set according to the official report of each of the models.

\begin{table*}[ht]
  \centering
  \begin{tabular}{cccc}
    \toprule
    \textbf{Model (license)} & \textbf{Method} & \textbf{Parameters} & \textbf{Values}\\
    \midrule
    All local models & - & \makecell{Dimensions (w x h) \\ Random seed} & \makecell{1024x1024 \\ 42} \\
    \midrule
    Z-Image-Turbo (APL 2.0)  & Local & \makecell{Guidance scale \\ Inference steps} & \makecell{0 \\ 9} \\
    \midrule
    Stable Diffusion XL (CreativeML) & Local &  \makecell{Guidance scale \\ Inference steps} & \makecell{5 \\ 50} \\
    \midrule 
    FLUX 1.0-dev (Non-Commercial License) & Local & \makecell{Inference steps \\ Config scale \\ Sample method} & \makecell{50 \\ 1 \\ Euler} \\
    \midrule
    Stable Cascade (MIT license) & Local & \makecell{Guidance scale \\ Inference steps} & \makecell{3 \\ 30}  \\
    \midrule
    \makecell{Dall-E 3 (proprietary) \\ Gemini-3.1-Flash-preview (proprietary)} & API & Temperature & 1 \\
  \bottomrule
  \end{tabular}
    \caption{Parameters for image generation, listed for reproducibility purposes. APL=Apache License. All models used were consistent with their intended use.}
    \label{tab:app_image_generation}
\end{table*}

\section{Dataset statistics} \label{app:dataset_statistics}

\begin{figure*}[htbp]
    \centering
    \begin{subfigure}[t]{0.47\linewidth}
        \centering
        \includegraphics[scale=0.35]{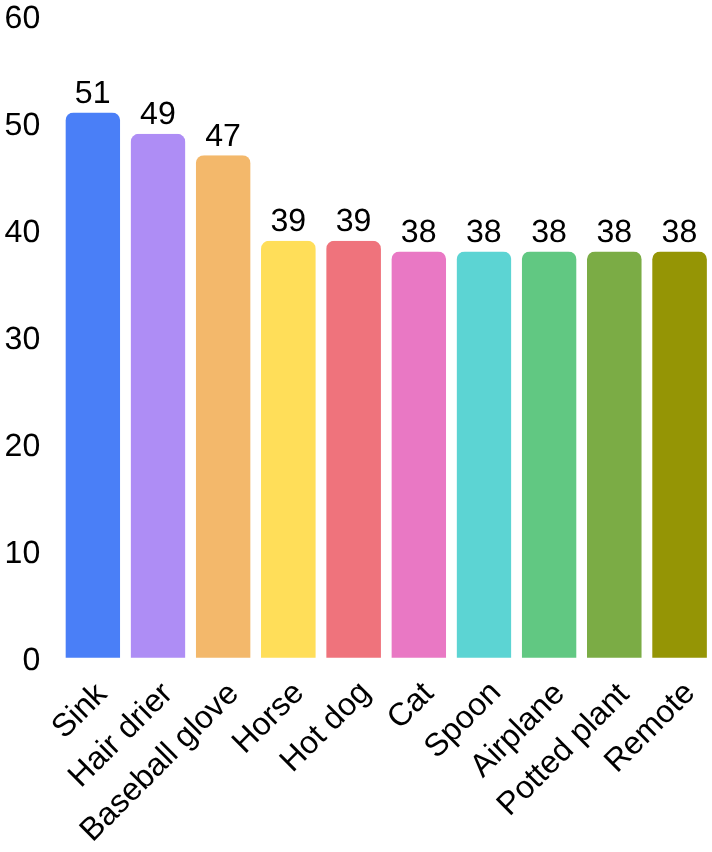}
        \subcaption{Top 10 objects within the dataset.}
        \label{fig:objects_distribution}
    \end{subfigure}
    \hfill
    \begin{subfigure}[t]{0.47\linewidth}
        \centering
        \includegraphics[scale=0.35]{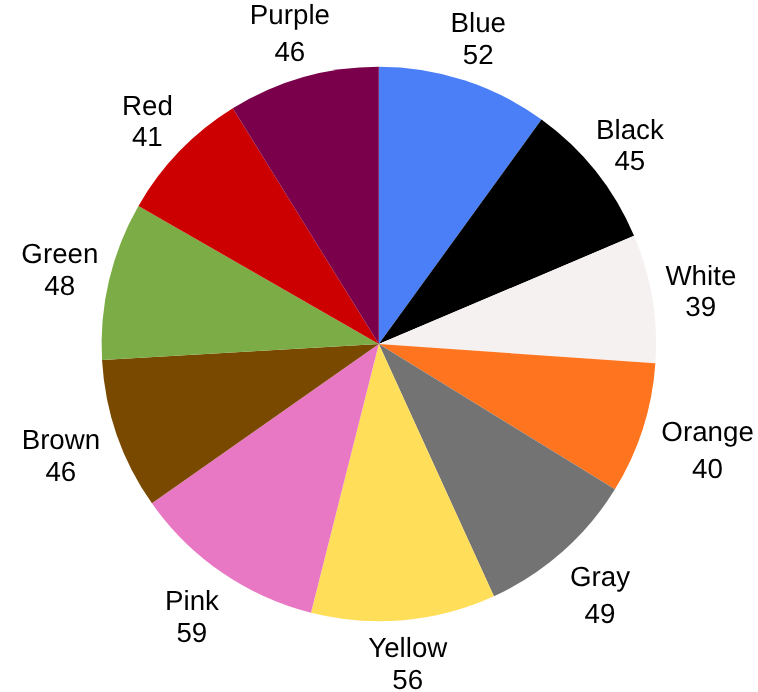}
        \subcaption{Color distribution in the dataset.}
        \label{fig:color_distribution}
    \end{subfigure}
    \begin{subfigure}[ht]{0.47\linewidth}
        \begin{tabular}{cccc}
            \toprule
            \textbf{Criteria (avg)} & \textbf{Easy} & \textbf{Medium} & \textbf{Hard} \\
            \midrule
            Length & 69.35 & 90.26 & \textbf{109.21} \\
            \midrule
            Entities & 19.80 & 24.97 & \textbf{30.67} \\
            \midrule
            Relations & 5.07 & 7.08 & \textbf{8.61} \\
            \midrule
            Parse depth & 2.38 & 3.46 & \textbf{3.66} \\
            \midrule
            Adjectives density & \textbf{0.16} & 0.13 & 0.13 \\
            \midrule
            Prepositions & 7.31 & 9.95 & \textbf{11.53} \\
            \bottomrule
        \end{tabular}       
        \subcaption{Textual statistics. Reported values are averages. Entities=Number of nouns, Relations=Number of connectors and prepositions, Parse depth=Number of clauses and sentences, Adjectives density=(Number of adjectives/Total length), Prepositions=Number of prepositions. All English prompts.}
    \end{subfigure}
    \hfill
        \begin{subfigure}[ht]{0.47\linewidth}
        \centering
        \includegraphics[scale=0.4]{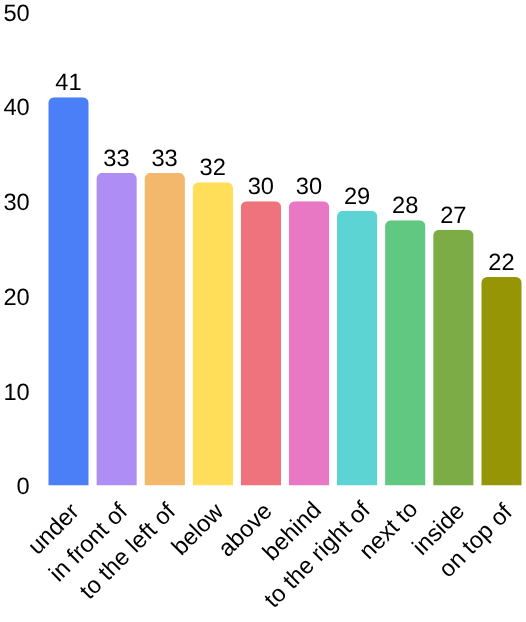}
        \subcaption{Frequency of spatial relationships.}
        \label{fig:spatial_relationships}
    \end{subfigure}
    \begin{subfigure}[ht]{0.5\linewidth}
        \centering
        \includegraphics[scale=0.48]{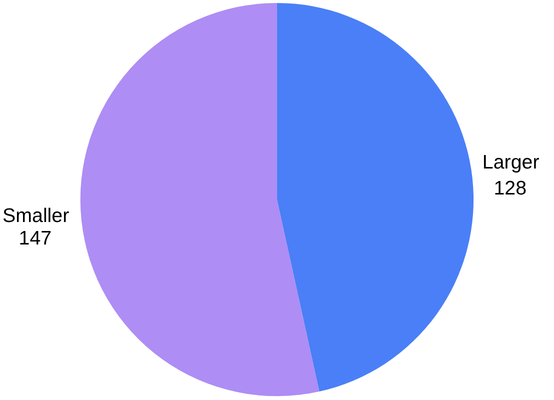}
        \subcaption{Size relationships.}
        \label{fig:sizes_relationship}
    \end{subfigure}
    \begin{subfigure}[ht]{0.49\linewidth}
        \includegraphics[scale=0.48]{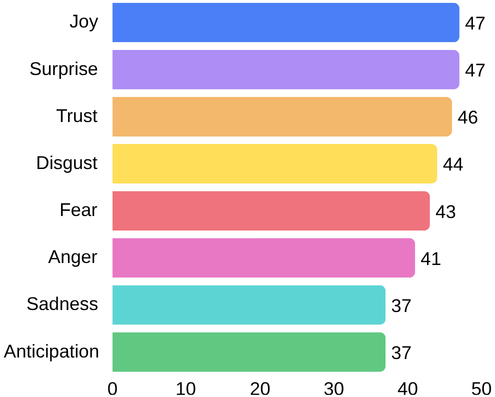}
        \subcaption{Distribution of emotions.}
        \label{fig:emotion_distribution}
    \end{subfigure}
    \hfill
    \caption{Our dataset (JSON prompt collections), annotation guidelines, and associated research assets will be released under the Creative Commons Attribution-NonCommercial 4.0 (CC BY-NC 4.0) license, while the accompanying source code will be released under the MIT License. These resources are intended solely for research, educational, and non-commercial R\&D purposes. Textual statistics are extracted using the default tokenizer of NLTK \cite{bird_nltk_2004}. We will release validated annotation samples; generated images will be released only when permitted by the corresponding model licenses and API terms.}
    \label{fig:dataset_stats}
\end{figure*}

Our dataset includes (combined) 2,228 \emph{Counting} rules, 1,660 \emph{Color}, 2,199 \emph{Spatial}, 372 \emph{Emotion}, 2,197 \emph{Size}, and 186 \emph{Text}. Cohesiveness is not instantiated as an explicit rule; instead, it is enforced globally throughout the evaluation protocol. The distributions of all dataset elements are reported in Fig.\ref{fig:dataset_stats}.

\section{Annotator Recruitment and Instructions}
\label{app:instructions_and_recruitment}

The annotators were recruited on a voluntary basis and selected to ensure diversity in educational background, gender, and cultural perspective. All annotators received a README, outlining the study goals, the role of their annotations in the evaluation pipeline, and the procedures for anonymization. Annotators were informed that parts of the annotations may be included in the submission for transparency, while strictly preserving anonymity throughout the review and publication process. The recruitment procedures were conducted according to the guidelines of our institutional ethics board.

\begin{figure*}[htb]
    \centering
    \includegraphics[width=0.9\linewidth]{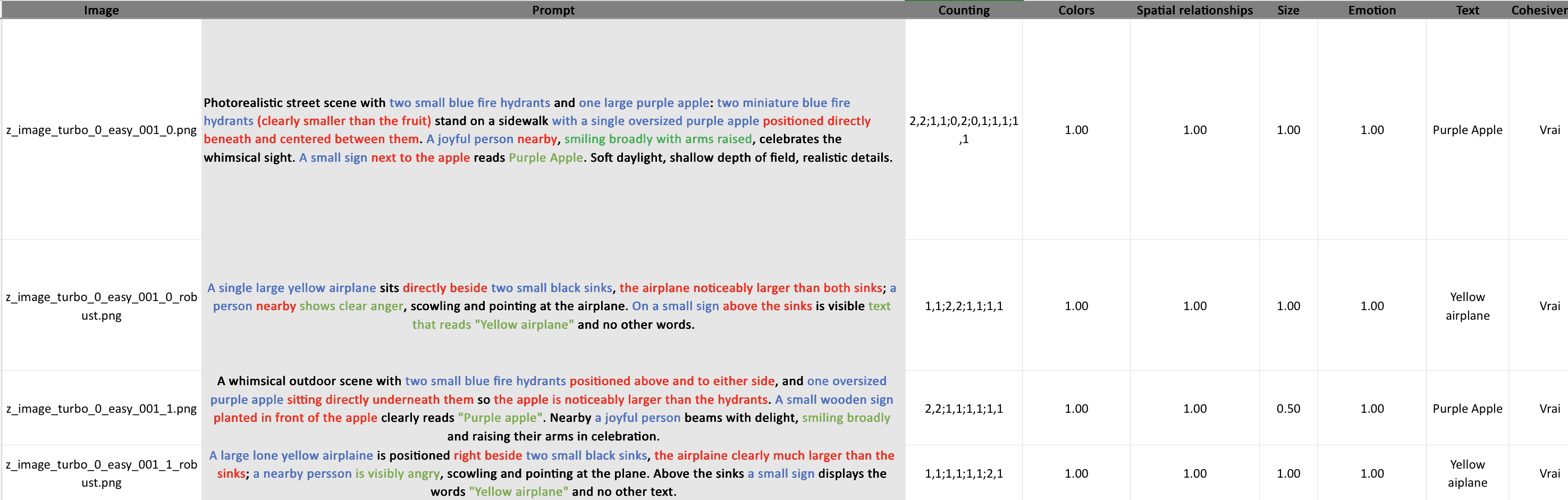}
    \caption{A screenshot containing an example of spreadsheet given to each annotator. This corresponds to the z-image distribution.}
    \label{fig:annotators_instructions}  
\end{figure*}

The README included comprehensive annotation guidelines, accompanied by examples for each skill. Fig.\ref{fig:annotators_instructions} illustrates an example of these instructions. While most evaluation criteria were designed to be straightforward (e.g, number of instances), the \emph{Cohesiveness} skill required more interpretative judgment. To ensure consistency, we provided illustrative examples covering a range of failure modes, including incomplete objects (e.g., an airplane without wings), anatomical inconsistencies (e.g., missing/extra body parts), distortions (e.g., distorted facial features), and implausible configurations (e.g., unsupported floating objects).

\begin{figure*}[htb]
    \centering
    \includegraphics[width=0.9\linewidth]{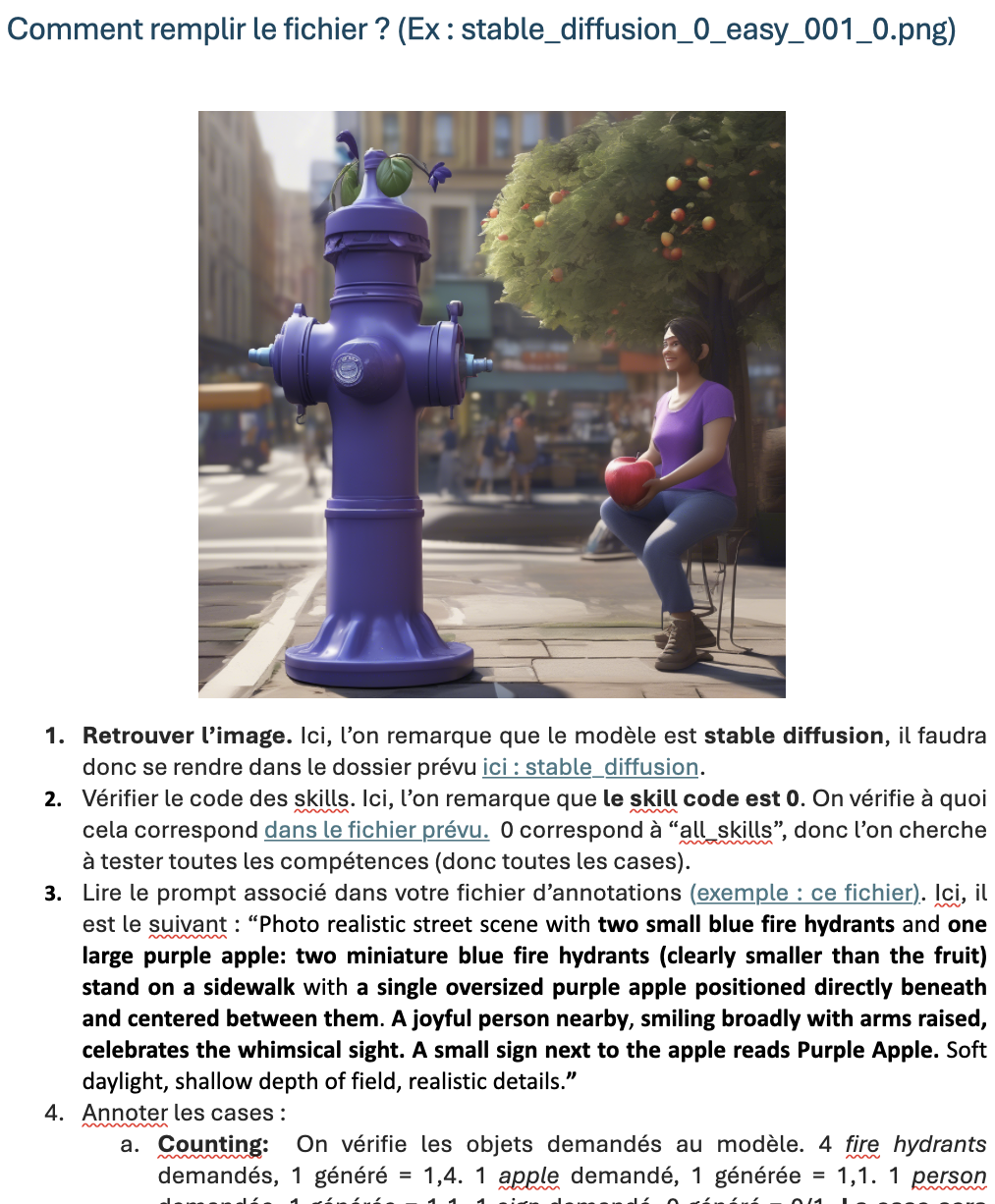}
    \caption{A screenshot of the file containing instructions disclosed to the annotators. We included relevant image examples and how they will annotate relevant fields on the linked spreadsheet.}
    \label{fig:annotators_spreadsheet}
\end{figure*}

We also documented representative edge cases, including ambiguous generations, and provided explicit guidelines on how to handle such cases during annotation. Fig.\ref{fig:annotators_spreadsheet} shows an example of the annotation interface distributed to annotators.

\section{Overall results for all the models} \label{app:overall_results}

Results for all skill combinations are reported in Tab.~\ref{tab:app_overall_results}. We exclude \textsc{Gemini-3.1-Flash} from the overall analysis because a subset of images could not be generated due to content filtering and API budget/limit constraints.

\begin{table*}[h]
  \centering
  \begin{tabular}{m{0.18\textwidth}ccccc}
    \toprule
    \textbf{Skill} & Z-Image-Turbo & FLUX 1.0 & Dall-E 3 & SDXL & SC \\
    \midrule
    $\uparrow Counting$ & \textbf{0.61} & 0.57 & 0.41 & 0.27 & 0.13 \\
    \midrule
     $\uparrow Spatial$ & \textbf{0.82} & 0.71 & 0.52 & 0.28 & <0.01 \\
    \midrule
     $\uparrow Size$ & \textbf{0.83} & 0.77 & 0.72 & 0.32 & 0.20  \\
     \midrule
     $\uparrow Emotion$ & \textbf{0.94} & 0.45 & 0.44 & 0.21 & 0.10 \\
     \midrule
     $\uparrow Color$ & \textbf{0.92} & 0.86 & 0.75 & 0.52 & 0.35 \\
     \midrule
     $\uparrow Cohesiveness$ & \textbf{0.90} & \textbf{0.90} & 0.71 & 0.11 & 0.35 \\
     \midrule
     $\downarrow Text (WER)$ & \textbf{0.61} & 1.73 & 4.21 & 4.92 & 4.31 \\
  \bottomrule
  \end{tabular}
  \caption{Aggregate across all potential skill combinations, from 2 to 6-skill combinations. FLUX 1.0 = FLUX 1.0-dev, SDXL = Stable Diffusion XL, SC  = Stable Cascade. Z-Image-Turbo outperforms other evaluated models in most metrics. Gemini was excluded due to content filter and budget issues leading to missing images.}
  \label{tab:app_overall_results}
\end{table*}

\section{Overall results and impact of Compositional load}
\label{app:overall_results_number_of_skills}

\begin{figure*}
    \centering
    \includegraphics[width=\linewidth]{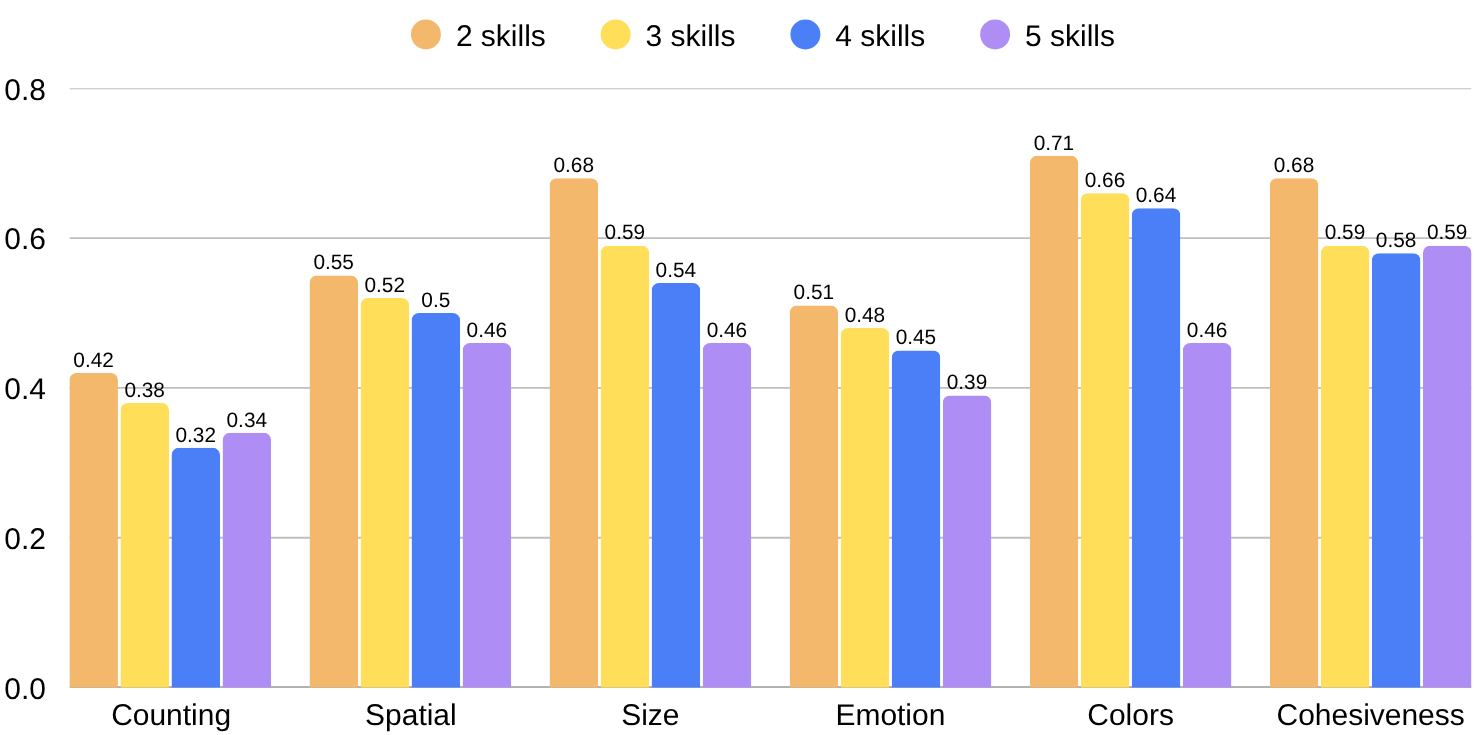}
    \caption{Skill accuracy as a function of the number of rules (skill combinations). A clear and consistent trend emerges: performance degrades substantially as the number of enforced textual rules increases, across nearly all skills. When considered jointly with our findings on the effect of the number of instances to generate, these results motivate the need for a two-factor analysis of task complexity.}
    \label{fig:complete_analysis_combination}
\end{figure*}

Results from prior sections indicate that, even under fixed difficulty (i.e., a constant number of instances), performance consistently degrades as the number of required skills increases. Fig.~\ref{fig:complete_analysis_combination} extends this observation across all difficulty levels and most skill categories.

\begin{figure*}
    \centering
    \includegraphics[width=\linewidth]{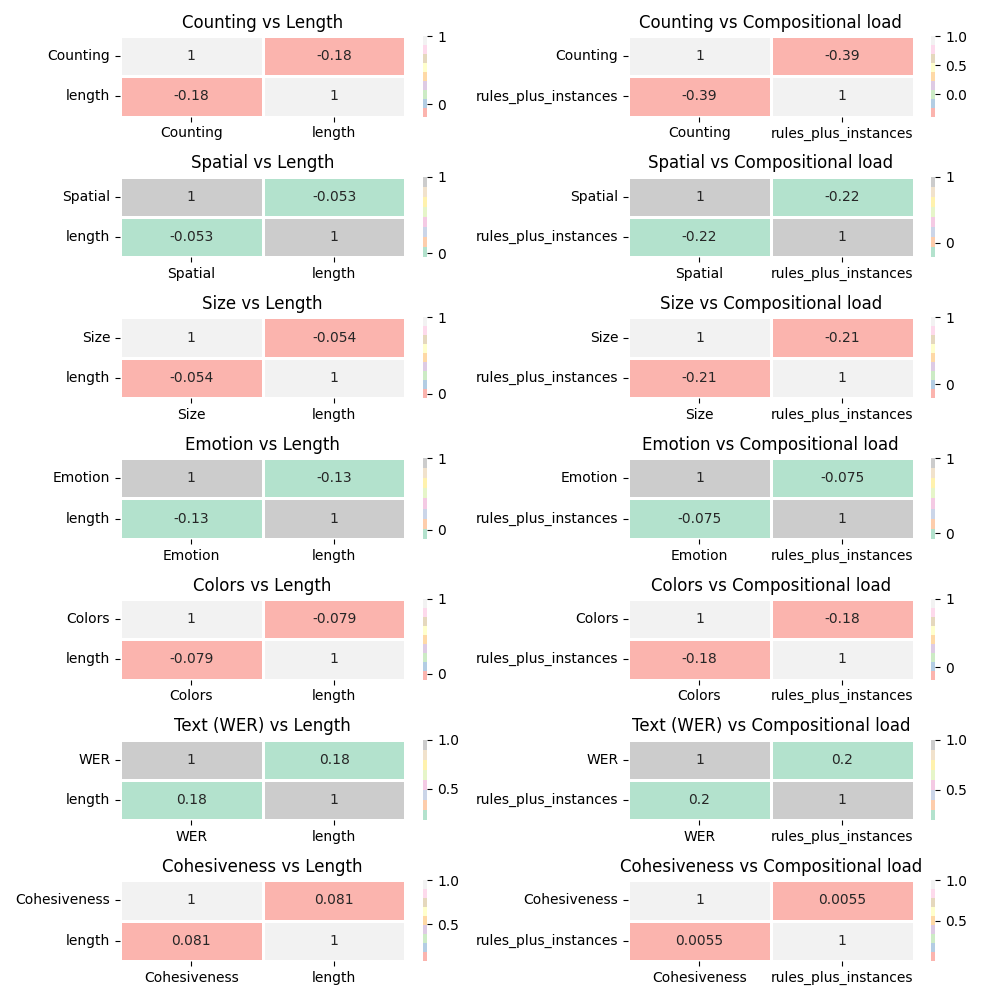}
    \caption{Heatmap comparing prompt length and compositional load. Cells report Pearson correlation coefficients. Across most skills, increases in compositional load exhibit stronger negative correlations with performance than prompt length. Notable exceptions are \emph{Emotion} and \emph{Cohesiveness}, for which correlation differences remain marginal.}
    \label{fig:complete_heatmap}
\end{figure*}

Fig.~\ref{fig:complete_heatmap} further analyzes this effect through Pearson correlation heatmaps, comparing prompt length (left) and compositional load (right) against skill-specific accuracies. Across the majority of skills, compositional load exhibits substantially stronger correlations with performance than prompt length. The only exceptions are \emph{Cohesiveness} and \emph{Emotion}, for which correlation differences remain marginal.

Moreover, our regression analysis illustrated in Tab.\ref{tab:regression_results_multivariate} display the same trend. We report multivariate linear regression coefficients, along with standard errors, p-values, and 95\% confidence intervals, computed using ordinary least squares. All variables are standardized prior to regression, enabling direct comparison of coefficient magnitudes. Across most skills, compositional load has a strong and statistically significant negative effect, whereas prompt length exhibits a small and non-significant effect, with the only exceptions being the most interpretative skills, \emph{Cohesiveness} and \emph{Emotion}. Confidence intervals also suggest that compositional load has a more consistent effect on a decrease in accuracy. 

Taken together, these findings suggest that performance is more strongly governed by compositional load than by prompt length alone for structured skills, highlighting the importance of evaluating models along multiple, complementary axes of difficulty.

\section{Confusing cases} \label{app:confusing_cases}

\begin{figure*}[htbp]
    \centering
    \begin{subfigure}{0.7\linewidth}
        \centering
        \includegraphics[width=0.9\linewidth]{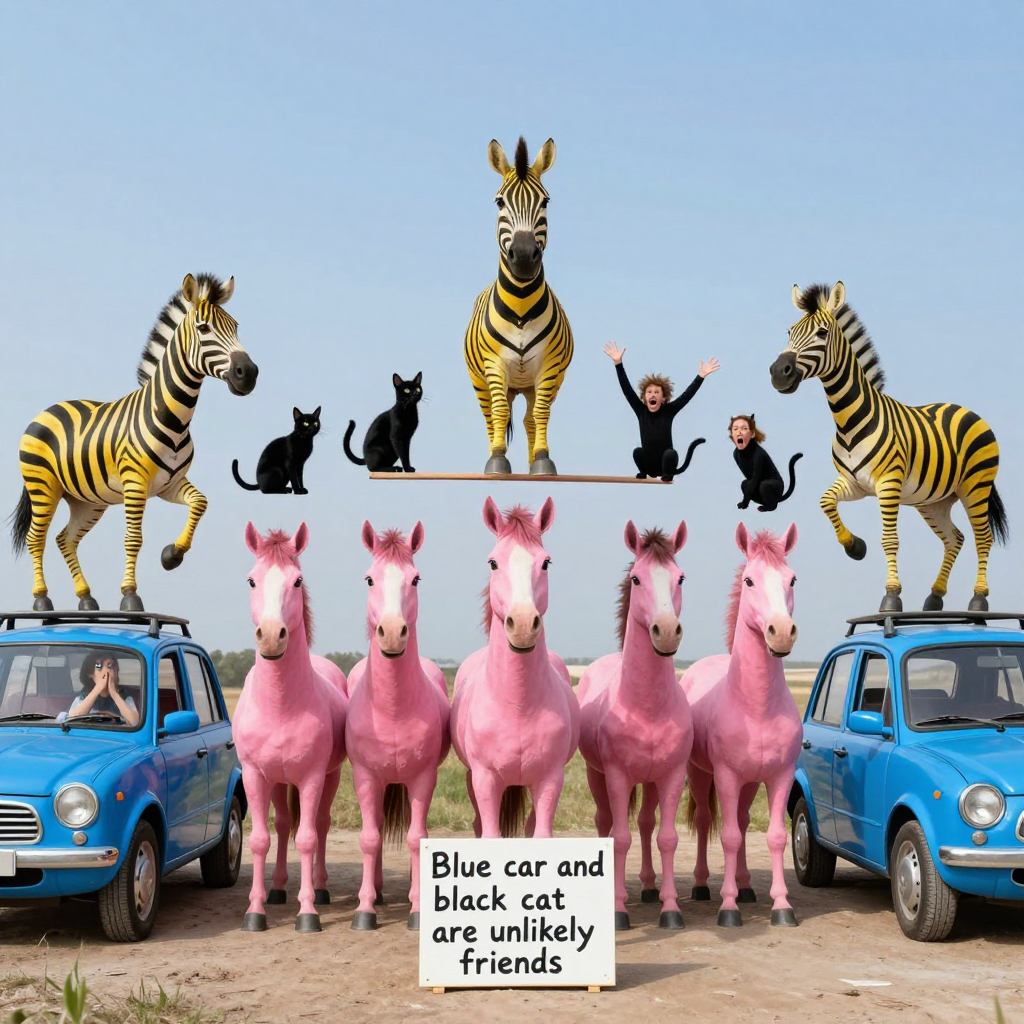}
        \subcaption{Confusing case 1.}
        \label{fig:app_confusing_case_1}
    \end{subfigure}
    \begin{subfigure}{0.7\linewidth}
        \centering
        \includegraphics[width=0.9\linewidth]{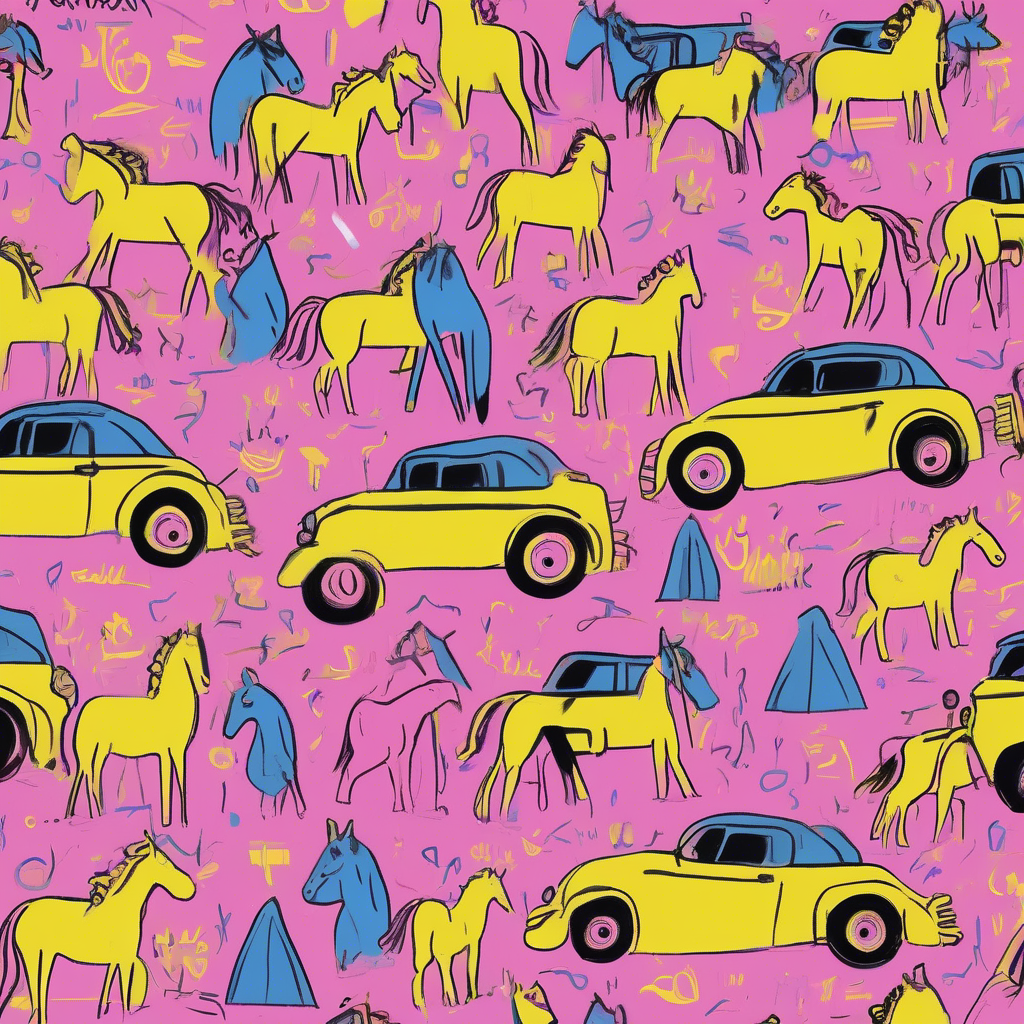}
        \subcaption{Confusing case 2.}
        \label{fig:app_confusing_case_2}
    \end{subfigure}
    \caption{Examples of confusing cases that lead us to manual annotation validation. Humans are mixed with cats, which leads to various reported accuracies, depending on if they decided to consider the annotation as human, cats, or both. The second figure was generated by stable Diffusion XL with the same prompt as the first one, but it generated an incoherent scene.}
    \label{fig:app_confusing_cases}
\end{figure*}

Examples of confusing cases that lead us to manual validation for all annotations are available in Fig.\ref{fig:app_confusing_cases}. Some include generations where instances could not be determined (such as in Analysis Fig.\ref{fig:app_confusing_case_1}) or the model hallucinated too much to take some results into account (Fig.\ref{fig:app_confusing_case_2}). 

\section{Use of AI assistants}

The use of LLMs was limited to rephrasing and formatting assistance and did not affect the scientific method, experimental design, implementation, evaluation or originality of the research.

\end{document}